%% file: 0-main.tex
\documentclass[twoside,11pt]{article}
\usepackage{xurl}
\usepackage{jair,theapa,rawfonts}

\usepackage[utf8]{inputenc} 
\usepackage[T1]{fontenc}

\usepackage{latexsym}
\usepackage{graphicx}
\usepackage{amsmath}
\usepackage{xspace}
\usepackage{subfig}
\usepackage{booktabs}
\usepackage{float}
\usepackage{cuted}
\usepackage{microtype}

\usepackage{url}
\usepackage{multirow, makecell}
\usepackage{tabularx}
\usepackage{comment}
\usepackage{adjustbox}
\usepackage{xcolor, colortbl}
\usepackage{booktabs}
\definecolor{Gray}{gray}{0.9}
\usepackage{siunitx}

\usepackage{todonotes}
\makeatletter
\newcommand*\iftodonotes{\if@todonotes@disabled\expandafter\@secondoftwo\else\expandafter\@firstoftwo\fi}
\makeatother

\newcommand{\genie}[1]{\textcolor{blue}{#1}}

\newcommand{\tod}{{\textsc{ToD}}\xspace}

\title{\textit{Crossing the Conversational Chasm:} \\ 
A Primer on Natural Language Processing for \\ Multilingual Task-Oriented Dialogue Systems}

\author{\name Evgeniia Razumovskaia \email er563@cam.ac.uk \\
\addr Language Technology Lab, University of Cambridge, UK \AND
\name Goran Glava\v{s} \email goran.glavas@uni-wuerzburg.de \\
\addr Center for AI and Data Science (CAIDAS), University of Würzburg, Germany \AND
\name Olga Majewska \email om304@cam.ac.uk \\
\addr Language Technology Lab, University of Cambridge, UK \AND
\name Edoardo M. Ponti \email edoardo-maria.ponti@mila.quebec \\
\addr Mila - Quebec AI Institute and McGill University, Canada \\ Language Technology Lab, University of Cambridge, UK \AND
\name Anna Korhonen \email alk23@cam.ac.uk \\
\addr Language Technology Lab, University of Cambridge, UK \AND
\name Ivan Vuli\'{c} \email iv250@cam.ac.uk \\
\addr Language Technology Lab, University of Cambridge, UK \\
PolyAI Limited, UK
}

\date{}

\begin{document}
\maketitle

\begin{abstract}
In task-oriented dialogue (\tod), a user holds a conversation with an artificial agent with the aim of completing a concrete task. Although this technology represents one of the central objectives of AI and has been the focus of ever more intense research and development efforts, it is currently limited to a few narrow domains (e.g., food ordering, ticket booking) and a handful of languages (e.g., English, Chinese). This work provides an extensive overview of existing methods and resources in multilingual \tod as an entry point to this exciting and emerging field. We find that the most critical factor preventing the creation of truly multilingual \tod systems is the lack of datasets in most languages for both training and evaluation. In fact, acquiring annotations or human feedback for each component of modular systems or for data-hungry end-to-end systems is expensive and tedious. Hence, state-of-the-art approaches to multilingual \tod mostly rely on (zero- or few-shot) cross-lingual transfer from resource-rich languages (almost exclusively English), either by means of (i) machine translation or (ii) multilingual representations. These approaches are currently viable only for typologically similar languages and languages with parallel / monolingual corpora available. On the other hand, their effectiveness beyond these boundaries is doubtful or hard to assess due to the lack of linguistically diverse benchmarks (especially for natural language generation and end-to-end evaluation). To overcome this limitation, we draw parallels between components of the \tod pipeline and other NLP tasks, which can inspire solutions for learning in low-resource scenarios. Finally, we list additional challenges that multilinguality poses for related areas (such as speech, fluency in generated text, and human-centred evaluation), and indicate future directions that hold promise to further expand language coverage and dialogue capabilities of current \tod systems.

\end{abstract}


\input{1-intro}

\input{2-to_dialog}
\input{3-current}
\input{4-challenges}

\input{6-conclusion}

\section*{Acknowledgments}
Evgeniia Razumovskaia is supported by a scholarship from Huawei. Ivan Vuli\'{c}, Olga Majewska, Edoardo M. Ponti, and Anna Korhonen are supported by the ERC Consolidator Grant LEXICAL: Lexical Acquisition Across Languages (no. 648909), the ERC PoC Grant MultiConvAI: Enabling Multilingual Conversational AI (no. 957356), and a Huawei research donation. Goran Glava\v{s} is supported by the Multi2ConvAI Grant (Mehrsprachige und domänenübergreifende Conversational AI) of the Baden-Württemberg Ministry of Economy, Labor, and Housing.

\clearpage
\input{7-appendix}

\vskip 0.2in
\bibliography{anthology,acl2020}
\bibliographystyle{theapa}

\end{document}

%% file: 1-intro.tex
\section{Introduction and Motivation}
\label{sec: introduciton}

Endowing machines with the ability to intelligently converse with humans has been one of the fundamental objectives in the pursuit of artificial intelligence. As compelling as it is challenging, developing dialogue systems capable of satisfying the end user on a par with human--human interaction remains an elusive target. Narrower in scope than general-purpose conversational assistants, {\em task-oriented dialogue} (\tod) systems\footnote{
Note that throughout the paper the word `task' can refer to either 1) the goal of a dialogue, or 2) a discriminative or generative problem, since both usages are standard in the fields of dialogue and NLP, respectively. The intended use of the term should always be easy to disambiguate from the context.
} 
\shortcite{gupta2005t,bohus2009ravenclaw,young2013pomdp,muise2019planning} have attracted interest from the scientific community as well as businesses as a so-far more feasible application. In fact, this technology has the potential to help or altogether replace human operators in focused problems and areas such as restaurant booking \shortcite{kim2014rcube,henderson2019polyresponse}, banking \shortcite{hardy-etal-2004-data,altinok2018ontology}, travel \shortcite{li2018microsoft,zang2020multiwoz}, or (non-emergency) healthcare \shortcite{laranjo2018conversational,denecke2019intelligent}.

The accelerated pace at which new milestones are reached across natural language applications thanks to the growing viability of deep learning techniques has recently catalysed dialogue-oriented research \shortcite[\textit{inter alia}]{ren-etal-2018-towards,wen-etal-2019-data,henderson2019convert,wu2020tod}. Coupled with the proliferation of affordable voice technology (e.g., Amazon Alexa, Google Assistant, Microsoft Cortana, Samsung Bixby), the so-far distant prospect of virtual assistants becoming part of everyday reality seems more attainable than ever. And yet, the momentum of developments in this area has mainly targeted a very small proportion of their potential beneficiaries, further deepening the chasm between speakers of dominant and under-represented languages in their access to state-of-the-art language technology.\footnote{For example, Amazon Alexa, one of the most popular virtual personal assistants, currently supports only eight resource-rich languages: English, French, German, Hindi, Italian, Japanese, Brazilian Portuguese, and Spanish.} Extending the reach of conversational technology is crucial for the democratisation of human--machine communication. This endeavour requires developing approaches that generalise across diverse language varieties and linguistic phenomena, are robust to cross-cultural differences in dialogue behaviours, and efficiently capitalise on available training data, whose scarcity continues to hold back truly multilingual conversational AI. 

In this survey, we take stock of the work carried out to date on multilingual \tod, discuss the main open challenges and lay out possible avenues for future developments. In particular, we aim to systematise the current research and know-how related to multilingual \tod, and shed new light on the following topics:
\vspace{1.2mm}

\noindent (Q1) Which \tod \textbf{datasets} are available in one or more languages other than English? What are their strengths and weaknesses?
\vspace{1.1mm}

\noindent (Q2) What are the best \textbf{methods}  and practices to incorporate language-specific information and conduct target language adaptation for multilingual and cross-lingual \tod?
\vspace{1.1mm}

\noindent (Q3) How can multilingual \tod take inspiration from other \textbf{related fields} of NLP research to better tackle low-resource scenarios? 
\vspace{1.1mm}

\noindent (Q4) What are the \textbf{future challenges} faced when developing \tod systems in several different languages, especially with respect to voice-based and human-centred dialogue?


\vspace{1.2mm}
Despite recent positive trends, and a slowly but steadily growing body of work on creating multilingual \tod data and methodology, 
our survey suggests that the pace of multilingual \tod research still lags behind other cross-lingual NLP work and other NLP tasks and applications (e.g., named entity recognition, dependency parsing, QA) with respect to linguistic diversity, training and evaluation data availability, cross-lingual transfer methodology, and joint multilingual modelling \shortcite{ponti2019modeling,Hedderich:2021naacl}. We hope that this survey will inspire more work in these areas. First, by drawing direct links (including similarities and differences) between \tod components and other cross-lingual NLP tasks, we advocate for the use and adaptation of existing techniques in support of multilingual \tod. Secondly, by surveying existing (multilingual) \tod resources and, consequently, exposing the current lack of training and evaluation resources for a large number of languages and domains, we emphasise the need for their creation.

%% file: 2-to_dialog.tex
\begin{figure}[!t]
\centering
\includegraphics[width=0.85\linewidth]{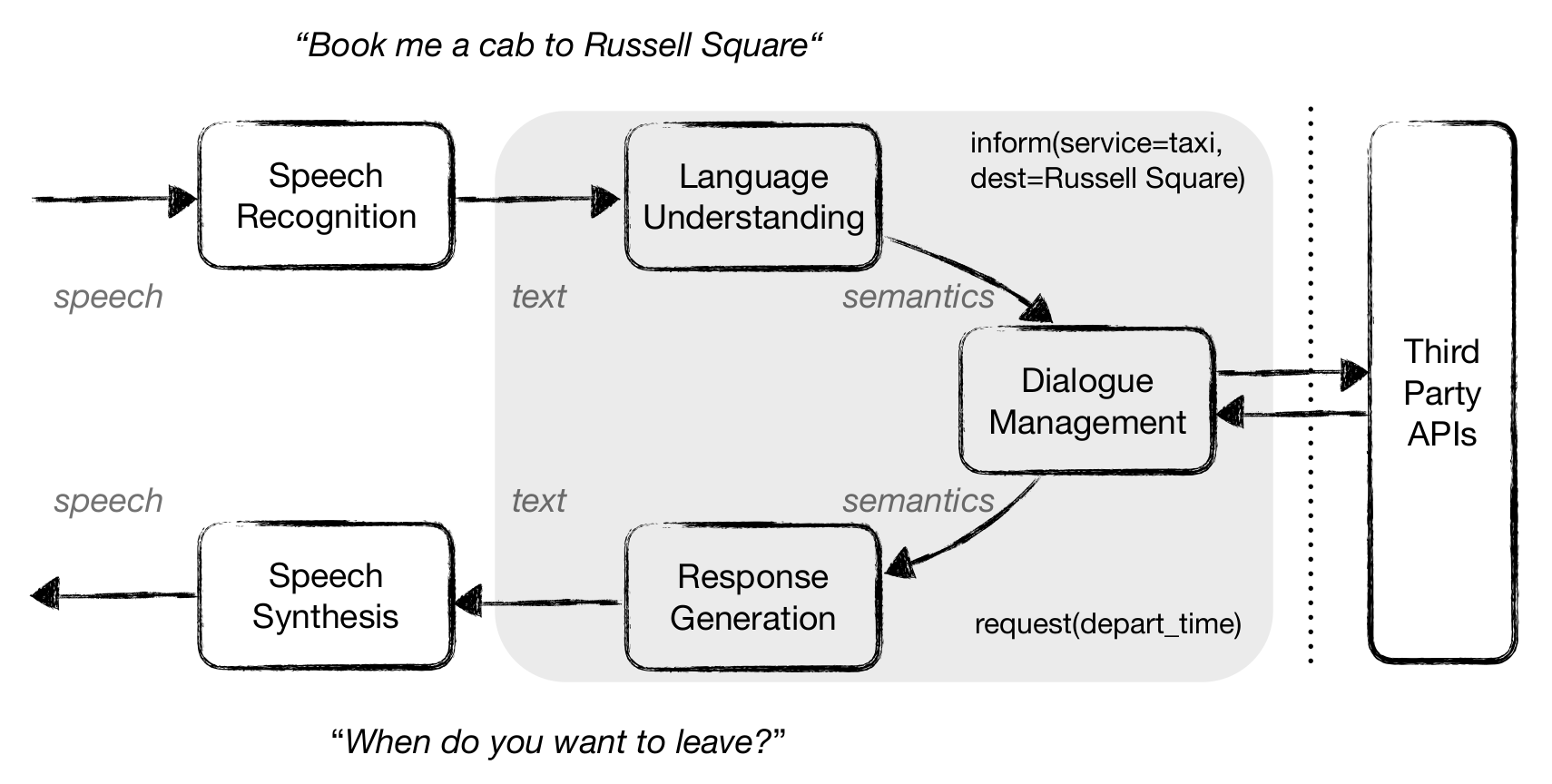}
\caption{The typical architecture of a modular dialogue system. The gray rectangle spans the modules operating on text, which are in the focus of this survey.}
\label{fig:modular}
\end{figure}

\section{Task-Oriented Dialogue Systems}
\label{sec: background}

The purpose of \tod systems, geared towards practical applications, is to assist users in completing a concrete task through conversational interaction \shortcite{Young:10b,Chen:2017,su2018deep}. Typically, the tasks are well-defined and the communication with the system has a binary outcome: either the task is successfully completed or not. Common examples of tasks include booking (restaurants, transportation, hotels), customer support (e.g., in banking or telecommunications), or retrieving and providing information (e.g., in healthcare or tourism).


For completeness, we first give a concise overview of the two existing approaches to task-oriented dialogue: (i) the \textit{modular} approach, in which \tod is broken down into a pipeline of sub-tasks and (ii) \textit{end-to-end} \tod, where a single neural model is trained to generate responses based on the context of the preceding interactions.

\subsection{Modular Task-Oriented Dialogue}
\label{sec:modular}

The modular approach to \tod addresses the complexity of goal-oriented dialogue by breaking it down into a sequence of sub-tasks. The solution, as depicted in Figure~\ref{fig:modular}, is a pipeline of independently trained and executed models (components): the output of each serves as the input to the next.\footnote{These outputs are often in a discrete and/or structured form that makes them non-differentiable with respect to the model parameters.} 
In this work, we focus our attention on dialogue systems that read text input and generate text output---which can be further extended to voice-based interactions by prepending an automatic speech recognition (ASR, speech-to-text) component to the start of the pipeline and appending a speech synthesis (TTS, text-to-speech) component to its end.\footnote{This extension, as discussed later in \S\ref{ss:speech}, comes with its own set of research challenges, but a comprehensive overview of (multilingual) ASR and TTS approaches falls way beyond the scope of this overview.} The three core text-based components of each modular \tod system are: natural language understanding (NLU), dialogue (policy) management (PM), and response generation (RG), outlined in what follows. 


\vspace{2mm}
\noindent \textbf{Natural Language Understanding (NLU).} In the context of \tod systems,\footnote{Note that, in the wider NLP context, NLU refers instead to the set of NLP tasks whose solution is presumed to require human-level competence in understanding natural language and composing meaning. Representative NLU tasks include natural language inference \shortcite{williams2018broad}, reading comprehension \shortcite{rajpurkar2016squad}, and commonsense reasoning \shortcite{sap2020commonsense}, among others.} natural language understanding consists of identifying the main goals and information expressed in the user's utterances. It usually encompasses two sub-tasks, namely \textit{intent classification} \shortcite<also known as \textit{dialogue act classification};>{ravuri2015recurrent,khanpour2016dialogue} and \textit{slot filling} \shortcite<also known as slot labelling or slot tagging;>{mesnil2014using,kurata2016leveraging}. The former is a single-label or multi-label classification task where one or more intent labels must be assigned to the whole utterance of a user; the latter, instead, requires to fill in predefined slots of information by extracting values from the content of the utterance. For example, the utterance ``\textit{Show me the flights from Boston to New York today}'' corresponds to the intent \texttt{find\_flight} and specifies values for three slots of information: \texttt{Boston} as departure location, \texttt{New York} as arrival location, and \texttt{today} as time. Given that the available slots depend on the utterance intent, the two tasks are often addressed jointly via multi-task learning \shortcite[\textit{inter alia}]{xu2013convolutional,guo2014joint,goo2018slot,chen2019bert,wu2020tod}. 




Traditionally, \tod systems included a component for \textit{dialogue state tracking} (DST), which falls in between NLU and dialogue management. The purpose of DST models \shortcite[\textit{inter alia}]{henderson2014word,mrkvsic2017neural,perez2017dialog,zhong2018global} is to maintain a \textit{dialogue belief state}, the summary of the dialogue history. This includes all the goals and slot values expressed by the user throughout the conversation. The input to DST at each user turn consists of the previous belief state and the outputs of the intent classification and slot filling modules; the output of DST is the new or updated belief state.
With the advent of attention-based Transformer models \shortcite{vaswani2017attention,devlin2019bert} and their ability to encode long sequences and capture long-distance dependencies, however, it has become possible to build latent representations of the dialogue history (from scratch) at every turn. Since maintaining an explicit belief state was no longer a strict requirement, this diminished the importance of DST in several Transformer-powered \tod systems \shortcite{wolf2019transfertransfo,budzianowski2019hello}. Nonetheless, more recent work demonstrates that explicit dialogue state tracking may still be beneficial even for these models \shortcite{lee2021improving,lee2021sumbt+}. Hence, we provide a brief overview of DST in multilingual \tod later in \S\ref{sec:stquo}.




\vspace{2mm}
\noindent \textbf{Dialogue (Policy) Management (PM)} refers to the \tod component responsible for choosing the system actions based on the current dialogue state. Approaches to PM can be broadly categorised into those based on rules, supervised learning, or reinforcement learning \shortcite<RL;>{su2018deep}. RL-based PM has been the predominant paradigm in recent years, because it is more flexible than rules and does not require utterance-level annotations like supervised learning. Nevertheless, a large number of conversations along with the final outcome label (i.e., successful or not successful) are still needed as a reward/penalty for RL. This has directed the research efforts towards simulating user interactions with the policy manager \shortcite{el2016sequence,cuayahuitl2017simpleds,cao2020adaptive}. PM models are independent from the dialogue language: they receive an abstract representation of the dialogue state from NLU and/or DST and produce an abstract action representation for the response generator; because of this, PM is not of particular interest in the context of multilingual \tod, as it inherits all the challenges and corresponding solutions directly from the research on monolingual PM.      




\vspace{2mm}
\noindent \textbf{Response Generation (RG)} is the module in charge of producing the system utterances in response to the user utterances, given the system action predicted by the policy manager. 
%
%
Much like PM, early work on RG relied on templates and rules hand-crafted by domain experts \shortcite[\textit{inter alia}]{langkilde-knight-1998-practical,stent-etal-2004-trainable,cheyer2006method,mirkovic2011dialogue}.
More recent data-driven approaches exploit ever-growing corpora of online human--human conversations (e.g., Reddit, Quora, Twitter) and produce system responses by either (1) \textit{generating} natural language utterances \shortcite<e.g.,>{sordoni-etal-2015-neural,li-etal-2016-deep,wen2017network,zhang2018generating,zhu-etal-2019-multi,peng-etal-2020-shot} or (2) \textit{retrieving} the most suitable response from a predefined set of candidate replies, also known as \textit{response selection} \shortcite<e.g.,>{lowe-etal-2017-towards,yang2018response,zhang-etal-2018-modeling,henderson-etal-2019-training}. 

Retrieval methods, on the one hand, offer the advantages of fluency, grammatical correctness, and high quality of the replies; modern neural natural language generation (NLG), in contrast, often produces overly general, incoherent, and grammatically erroneous utterances \shortcite{li-etal-2016-diversity,gao-etal-2018-neural,serban-etal-2016-generating}. On the other hand, reliance on fixed lists of candidate responses constrains the system versatility, making response quality highly dependent on the size of the response inventory (based on a corpus of human--human interactions). Hybrid methods combine the best of both worlds \shortcite{song2018ensemble,weston-etal-2018-retrieve,pandey-etal-2018-exemplar,yang2019hybrid}: 
they first retrieve a set of response candidates and then provide them, together with the user utterance (or wider dialogue history), as input to a generative model, which then produces the final system response.


\subsection{End-to-end Task-Oriented Dialogue (e2e)} The components of a modular \tod system are trained and executed in isolation. As a consequence, the later pipeline components inherit the errors of earlier components, but cannot provide feedback to correct them. To remedy this well-known issue of pipeline learning systems, \tod can instead rely on end-to-end neural architectures \shortcite{wen2017network,liu2018dialogue,qin2020dynamic}. Some e2e models mirror the modules of the traditional pipeline \shortcite{wen2017network}, the parameters of which are all jointly tuned in a single training procedure; however, this is by no means indispensable.  

On the one hand, end-to-end training addresses the main issue of modular \tod, namely component isolation and error cascading. On the other hand, e2e models aim to capture complex interactions among intents, policies, and responses in a latent representation space: this typically requires a large number of model parameters, whose reliable estimation in turn requires a large number of examples. Because of this exigency, e2e models have been much more successful in open-domain conversations \shortcite<i.e., chatbots;>[inter alia]{serban2016building,lowe2017training,adiwardana2020towards,zhang2020dialogpt} than in \tod.         


\subsection{Why Is Developing Multilingual Dialogue Systems Difficult?}
\label{sec:gaps}

Sub-tasks of modular \tod systems can be interpreted as specific instances of more general classes of NLP problems. For instance, intent classification is a short-text classification task, whereas slot-filling can be cast as a sequence labelling, span extraction, or even a question answering task (cf.\S\ref{ss:lowres}). 
Top performance in such tasks is obtained with supervised learning. Adopting this strategy for multilingual \tod, however, postulates the availability of labelled data for most natural languages, for each task of interest. The impossibility of fulfilling this precondition, due to the cost- and time-intensive nature of the annotation process, is the principal bottleneck of multilingual NLP \shortcite{Joshi:2020acl}: most existing datasets for general language understanding and reasoning tasks \shortcite{conneau2018xnli,hu2020xtreme,ponti2020xcopa} have training portions only in English. The fact that \tod most commonly entails a pipeline of supervised models makes this prospect even more remote: for optimal \tod in a given language, one would need to collect language-specific annotations for each sub-task (intent detection, slot filling, response selection and/or response generation).    

The absence of language-specific annotations for most languages steered research efforts towards cross-lingual transfer, where knowledge from resource-rich languages is harnessed to facilitate zero-shot or few-shot predictions in resource-lean languages. 
Cross-lingual transfer, however, requires bridging between languages whose properties may vary considerably. Therefore, it is more likely to succeed when the source and target languages are close in terms of typology, family, and/or area \shortcite{lin2019choosing,lauscher2020zero}. Recently,
cross-lingual word embeddings \shortcite{Ruder:2019jair,glavavs2019properly} and  massively multilingual pre-trained encoders (MEs) based on Transformers \shortcite{devlin2019bert,Conneau:2020acl} have been the most popular vehicles for cross-lingual transfer of NLP models. 
While the transfer capabilities of MEs were initially showcased as remarkable \shortcite{pires2019multilingual,wu2019beto}, \shortciteA{lauscher2020zero} and \shortciteA{Wu:2020repl} have recently demonstrated that their effectiveness is drastically diminished in distant target languages whose monolingual corpora are small-sized. 
Likewise, MEs might represent a viable solution for task-oriented dialogue only in well-documented languages similar to English.

An alternative paradigm for cross-lingual transfer in NLP is contingent on (neural) machine translation \shortcite<MT;>{banea-etal-2008-multilingual,durrett-etal-2012-syntactic,conneau2018xnli} and comes in two flavours: `translate test' maps the evaluation data onto the source language, whereas `translate train' maps the training data onto the target language. In both cases, a \textit{monolingual} model (in the source or target language, respectively) can be subsequently deployed on the task. As a consequence, this approach is not prone to the `curse of multilinguality' that plagues massively multilingual encoders \shortcite{Arivazhagan:2019arxiv,Conneau:2020acl}. On the other hand, the coverage of translation-based transfer is constrained in terms of languages, as the development of MT models hinges upon the availability of parallel corpora, and in terms of tasks, as only sentence-level (but not token-level) properties can be preserved in translation. In modular \tod, this approach is therefore compatible with a subset of NLU tasks (intent classification and DST, but not slot filling) and RG. In end-to-end \tod, it could instead be operationalised by integrating MT modules into the pipeline before NLU and after RG. Foreshadowing \S\ref{subsec:e2e}, however, this idea has never been given concrete form thus far.

Although cross-lingual transfer in \tod is conceptually sound and practically feasible, there is only anecdotal evidence for its wide-range effectiveness \shortcite{schuster2019cross,liu2019zero}, primarily due to the scarcity of {multilingual evaluation benchmarks}. 
For the same reason, it also remains unclear how different transfer approaches (e.g., machine translation and multilingual encoders) compare against each other, and which one should be preferred in relation to particular sets of \tod-related tasks, languages, and domains.

%


%% file: 3-current.tex
\section{Existing Methods and Resources for Multilingual \tod}
\label{sec:stquo}

\begin{figure}[]
    \centering
    \includegraphics[width=0.8\textwidth]{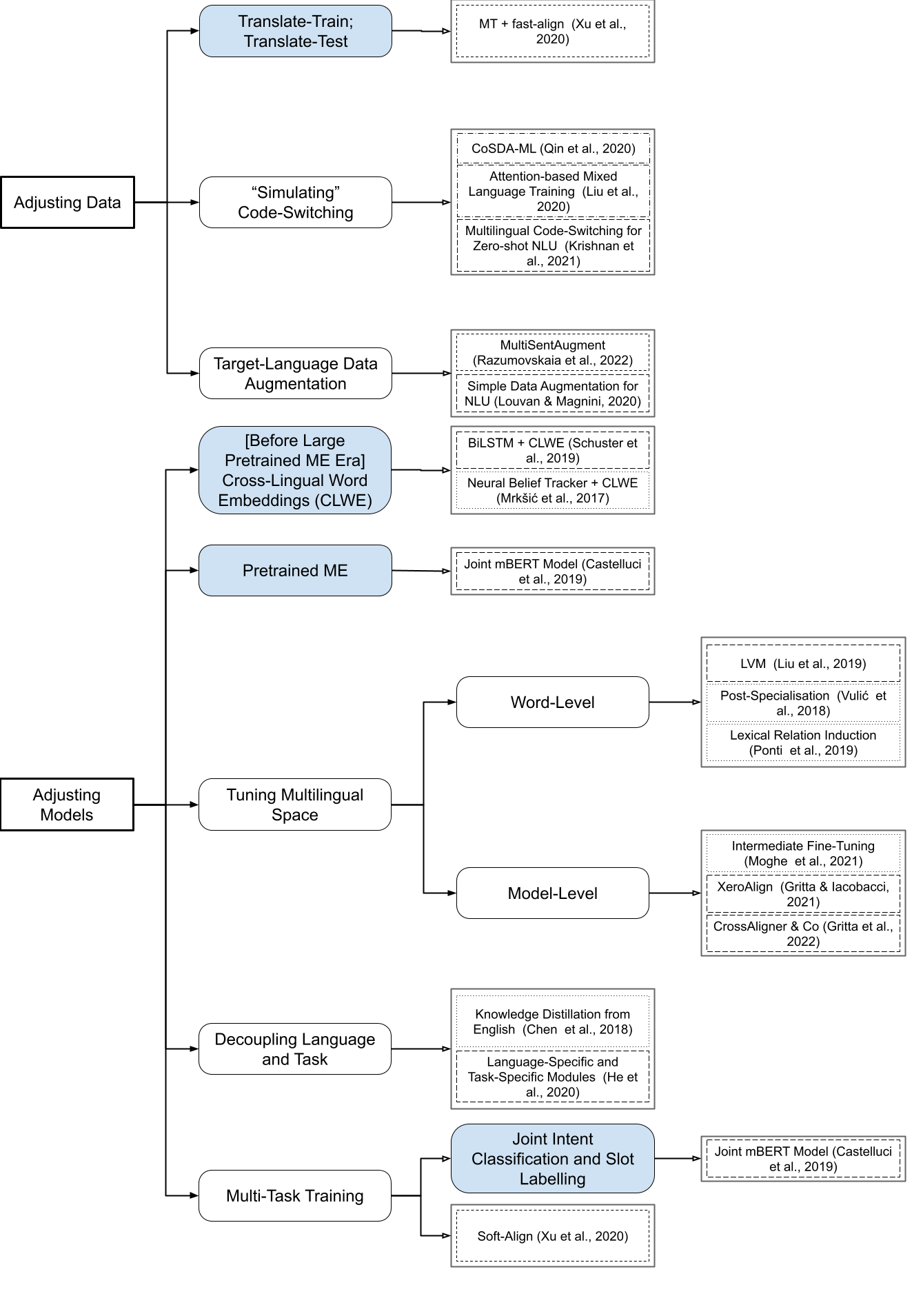}
    \caption{
    Taxonomy of approaches to multilingual dialogue systems, with prominent examples for established methods. Abbreviations: ME -- multilingual encoders.  
    The shaded nodes indicate well established methods, commonly used as baselines; for these methods we provide example papers where they were used. Nodes with dashed edges denote methods that are commonly applied to NLU tasks; nodes with dotted edges indicate methods specifically designed for DST; The nodes with mixed dashes-and-dots edges correspond to methods that can be effectively used for both tasks.}
    \label{fig:taxonomy}
\end{figure}

We now provide an overview of existing methods and resources for multilingual and cross-lingual \tod. Fig. \ref{fig:taxonomy} displays the taxonomy of these approaches. We group the approaches to multilingual and cross-lingual \tod in two main categories: (a) those that modify (i.e., adjust) the data, either training or evaluation data, in order to achieve better multilingual or cross-lingual transfer performance; and (b) those that introduce models (or adjustments to existing models) better tailored for multilingual \tod or cross-lingual \tod transfer in general, or a particular target setup (particular language(s) and/or task(s)).    
%
%
We next cover the multilingual and cross-lingual work focusing on each component of modular \tod (\S\ref{subsec:nlu}--\ref{subsec:nlg}), and then follow with the overview of the multilingual and cross-lingual efforts in end-to-end \tod (\S\ref{subsec:e2e}).


\subsection{Natural Language Understanding (NLU)}
\label{subsec:nlu}

\noindent \textbf{Joint versus Separate Training.} NLU approaches can be divided into two groups depending on whether they tackle intent classification and slot filling (i) jointly, in multi-task training regimes \shortcite[inter alia]{schuster2019cross,liu2019zero,xu2020multiatis++,bunk2020diet} or (ii) independently, addressing each of the tasks with separately trained models \shortcite[\textit{inter alia}]{ren2020intentiondetectionsiamese,he2020multilevelcrosslingslotpred,arora2020cross}. Joint multi-task training, besides potentially reducing the number of parameters, is advantageous for NLU \shortcite{zhang2019jointlearningwithbert}, as the two tasks are clearly interdependent: intuitively, the slots that may be filled with values given an utterance also depend on the intent of that utterance. 

\vspace{2mm}
\noindent \textbf{Cross-lingual Transfer Methods.} Given the absence of sufficiently large training data in many languages, the default approach to multilingual NLU is (zero-shot or few-shot) transfer of models trained on English datasets by means of massively multilingual Transformer-based encoders \shortcite{zhang2019jointlearningwithbert,xu2020multiatis++,siddhant2020evaluating,krishnan2021multilingual}. While most of the work relies on MEs pretrained via masked language modelling, such as mBERT \shortcite{devlin2019bert} and XLM-R \shortcite{Conneau:2020acl}, \shortciteA{siddhant2020evaluating} show that MEs pretrained via machine translation leads to more effective zero-shot transfer for intent classification.
Moreover, \shortciteA{liu2021wordorder} illustrated how mBERT-based sequence labelling overfits to the word order of the source language. Regularising for word order information (e.g., by removing positional embeddings or by shuffling tokens) benefits cross-lingual transfer in various sequence labelling tasks, including \tod slot filling. \shortciteA{he2020multilevelcrosslingslotpred} demonstrated the efficiency of dividing slot tagging models into language-specific modules and modules shared across languages, and complementing the input data with language discriminators.

Furthermore, representation sub-spaces of individual languages in MEs can be better aligned with the explicit supervision of word or sentence alignments \shortcite{cao2020adaptive,Conneau:2020emerging}. 
\shortciteA{kulshreshtha2020crosslingualalignmentmethods} found that these yielded superior zero-shot performance
on three \tod slot-filling datasets \shortcite{upadhyay2018multiatis,schuster2019cross,bellomaria2019almawave} and five target languages (Hindi, Turkish, Spanish, Thai, and Italian) 
compared to vanilla MEs. Recent work \shortcite{gritta-iacobacci-2021-xeroalign,gritta2022crossaligner} shows that contrastive learning can significantly improve the cross-lingual alignment of text representations for \tod slot-filling and intent classification, even if starting from pretrained massively multilingual encoders, known to suffer from the ``curse of multilinguality'' \shortcite{Conneau:2020acl,pfeiffer2022lifting}.  

Earlier work leveraged static cross-lingual word embedding spaces \shortcite<CLWEs;>[\textit{inter alia}]{mikolov2013exploiting,smith2017offline,artetxe2018robust,joulin2018loss,patra-etal-2019-bilingual,glavavs2020non} as a mechanism for cross-lingual transfer of NLU models \shortcite{upadhyay2018multiatis,chen2018xlnbt,schuster2019cross}. 
In fact, even a small number of word alignments in the actual \tod domain of interest is sufficient to refine the CLWE space \shortcite{liu2019zero} or simulate code-switching data \shortcite{liu2019attention}. 


%
%


\begin{table}[!t]
\resizebox{\textwidth}{!}{
\begin{tabular}{llccccc} 
\toprule
\textbf{Dataset}                                          & \textbf{Task}                                                                       & \textbf{Language} & \textbf{Domains}                                                                                  & \textbf{Size} & \textbf{\# Intents} & \textbf{\# Slots}                                              \\ 
\midrule
\rowcolor{Gray}
\begin{tabular}[c]{@{}l@{}}MEDIA \\ \shortcite{bonneau2005mediacorpus}\end{tabular}          & slot extraction                                                                     & fr                & hotel reservations                                                                                & 15000         & N/A                 & 83                                                             \\ 
\hline
\begin{tabular}[c]{@{}l@{}}SLU-IT \\ \shortcite{castellucci2019slu-it} \end{tabular}         & \begin{tabular}[c]{@{}l@{}}intent classification;\\ slot extraction\end{tabular}    & it                & \begin{tabular}[c]{@{}c@{}}7 domains (e.g., \\ music, weather,\\ restaurant)\end{tabular} & 7142          & 7                   & 39                                                             \\ 
\hline
\begin{tabular}[c]{@{}l@{}}Almawave-SLU\\ \shortcite{bellomaria2019almawave} \end{tabular}    & \begin{tabular}[c]{@{}l@{}}intent classification;\\ slot extraction\end{tabular}    & it                & \begin{tabular}[c]{@{}c@{}}7 domains (e.g., \\ music, weather,\\ restaurant)\end{tabular} & 14484         & 7                   & 39                                                             \\ 
\hline
\rowcolor{Gray}
\shortciteA{zhang2017first}  & intent classification & zh                & \begin{tabular}[c]{@{}c@{}}chit chat; \\ task-oriented\end{tabular}                               & 4000          & 31                  & N/A                                                            \\ 
\hline
\rowcolor{Gray}
\begin{tabular}[c]{@{}l@{}}ECSA dataset \\ \shortcite{gong2019escadataset}\end{tabular}   & \begin{tabular}[c]{@{}l@{}}slot extraction; \\ named entity extraction\end{tabular} & zh                & online commerce                                                                                   & 27615         & N/A                 & \begin{tabular}[c]{@{}c@{}}N/A\\ (sequence tags)\end{tabular}  \\ 
\hline
\begin{tabular}[c]{@{}l@{}}Chinese ATIS\\ \shortcite{he2013chineseatis} \end{tabular}    & \begin{tabular}[c]{@{}l@{}}intent classification;\\ slot extraction\end{tabular}    & zh                & airline travels                                                                                   & 5871          & 21                  & 120                                                            \\ 
\hline
\begin{tabular}[c]{@{}l@{}}Vietnamese ATIS\\ \shortcite{dao2021vietnameseatis} \end{tabular} & \begin{tabular}[c]{@{}l@{}}intent classification;\\ slot extraction\end{tabular}    & vi                & airline travels                                                                                   & 5871          & 25                  & 120                                                            \\
\bottomrule
\end{tabular}
}
\caption{Monolingual NLU datasets in non-English languages. Shaded rows correspond to datasets created from scratch, white rows to datasets translated from English. A non-exhaustive list of English NLU datasets is provided in the Appendix.}
\label{table:nlu monolingual}
\end{table}

\begin{table}[!t]
\resizebox{\textwidth}{!}{
\begin{tabular}{llccccc} 
\toprule
\textbf{Dataset} & \textbf{Task(s)} & \textbf{Languages} & \textbf{Domain(s)} & \textbf{Size} & \textbf{\# intents} & \textbf{\# slots} \\ 
\midrule
\begin{tabular}[c]{@{}l@{}}Multilingual TOP\\ \shortcite{schuster2019cross}\end{tabular}                & \begin{tabular}[c]{@{}l@{}}intent classification;\\ slot extraction\end{tabular}                    & en, es, th & \begin{tabular}[c]{@{}c@{}}alarm, reminder, \\ weather\end{tabular}  & \begin{tabular}[c]{@{}c@{}}43323 [en]\\ 8643 [es]\\ 5082 [th]\end{tabular} & 12 & 11 \\ 
\hline
\begin{tabular}[c]{@{}l@{}}ATIS in Chinese \\ and Indonesian \\  \shortcite{susanto2017neural} \end{tabular} & \begin{tabular}[c]{@{}l@{}}semantic parsing;\\ slot extraction\end{tabular} & en, zh, id & airline travels & 5371  & N/A  & \begin{tabular}[c]{@{}c@{}}120\\ (166; \\ $\lambda$-calculus)\end{tabular}  \\ 
\hline
\begin{tabular}[c]{@{}l@{}}Multilingual ATIS\\  \shortcite{upadhyay2018multiatis} \end{tabular}               & \begin{tabular}[c]{@{}l@{}}intent classification;\\ slot extraction\end{tabular}                    & en, hi, tr & airline travels & \begin{tabular}[c]{@{}c@{}}1493 [hi]\\ 1315 [tr]\end{tabular}  & 21  & 120 \\ 
\hline
\begin{tabular}[c]{@{}l@{}}MultiATIS++\\ \shortcite{xu2020multiatis++} \end{tabular}& \begin{tabular}[c]{@{}l@{}}intent classification;\\ slot extraction\end{tabular} & \begin{tabular}[c]{@{}c@{}}en, es, pt, \\ de, fr, zh, \\ ja, hi, tr\end{tabular} & airline travels & \begin{tabular}[c]{@{}c@{}}5871 [en, es,\\pt, de, fr, ja\\zh]\\ 2493 [hi]\\ 1353 [tr]\end{tabular} & \begin{tabular}[c]{@{}c@{}}18;\\ 17 [hi],\\ 17 [tr]\end{tabular} & \begin{tabular}[c]{@{}c@{}}84\\ 75 [hi]\\ 71 [tr]\end{tabular}              \\ 
\hline
\begin{tabular}[c]{@{}l@{}}MTOP\\  \shortcite{li2020mtop} \end{tabular} & \begin{tabular}[c]{@{}l@{}}semantic parsing;\\intent classification;\\ slot extraction\end{tabular} & \begin{tabular}[c]{@{}c@{}}en, de, fr, \\ es, hi, th\end{tabular} & \begin{tabular}[c]{@{}c@{}}11 domains (e.g., \\ music, news, \\ recipes)\end{tabular} & \begin{tabular}[c]{@{}c@{}}18788 [en]\\ 16585 [de]\\ 15459 [fr]\\ 16182 [es]\\ 15195 [hi]\\ 18788 [th]\end{tabular} & 117 & 78 \\ 
\hline
\begin{tabular}[c]{@{}l@{}}xSID$^{\spadesuit}$\\ \shortcite{van-der-goot-etal-2021-masked}  \end{tabular}      &   \begin{tabular}[c]{@{}l@{}}intent classification;\\slot extraction\end{tabular}  &  \begin{tabular}[c]{@{}c@{}}ar, da, de, \\ de-st, en, \\ id, it, kk, \\ nl, sr, tr,\\ zh~\end{tabular} & \begin{tabular}[c]{@{}c@{}}6 domains, \\ combining  \\ Multilingual \\TOP and \\SNIPS\end{tabular} &\begin{tabular}[c]{@{}c@{}} 800 [ar, da, \\ de, de-st, en, \\id,  it, kk, nl, \\ sr, tr, zh]  \\ 400 [ja] \end{tabular} & 16                                                               & 33 \\ \hline
\begin{tabular}[c]{@{}l@{}}MASSIVE\\ \shortcite{bastianelli-etal-2020-slurp} \end{tabular}      &   \begin{tabular}[c]{@{}l@{}}intent classification;\\slot extraction\end{tabular}  &  \begin{tabular}[c]{@{}c@{}}51 languages,\\incl. am, fi, \\mn, my, nb,\\alb, tl, ur\end{tabular} & \begin{tabular}[c]{@{}c@{}}18 domains (e.g., \\ alarm, calendar, \\music, email)\end{tabular} &\begin{tabular}[c]{@{}c@{}} 19521[per lang]\end{tabular} & 60                                                               & 55 \\
\bottomrule
\end{tabular}
}
\caption{Multilingual NLU datasets. $^{\spadesuit}$xSID is an evaluation-only dataset, whereas all the other datasets listed here also contain target-language training sets.}
\label{table:nlu multilingual}
\end{table}



\vspace{2mm}
\noindent \textbf{Available Datasets.} 
While multilingual NLU is better resourced than other tasks in modular \tod, the landscape of existing datasets is still sparsely populated. Table~\ref{table:nlu monolingual} lists available monolingual NLU datasets in languages other than English and Table~\ref{table:nlu multilingual} lists multilingual NLU datasets.\footnote{For completeness, we also provide a (non-exhaustive) collection of English-only NLU resources in the Appendix.}   
Most of these datasets have been obtained by translating pre-existing or newly created examples from English: SNIPS \shortcite{coucke2018snips} was translated by \shortciteA{castellucci2019slu-it,bellomaria2019almawave} into Italian; ATIS was translated by \shortciteA{he2013chineseatis} into Chinese,
by \shortciteA{dao2021vietnameseatis} into Vietnamese, and by \shortciteA{susanto2017neural,upadhyay2018multiatis,xu2020multiatis++} into 10 typologically diverse languages. 
However, a small group of monolingual datasets \shortcite{bonneau2005mediacorpus,zhang2017first,gong2019escadataset} were built from scratch based on original conversations in the target language.


Ideally, NLU models should be able to generalise both over languages and over domains. Most existing datasets, however, either cover multiple domains in a single language \shortcite{hakkani2016multi,liu2019hwu64} or the same domain across different languages \shortcite{xu2020multiatis++}. Fortunately, the most recent generation of NLU datasets \shortcite{li2020mtop,van-der-goot-etal-2021-masked,majewska2022cross} is both multi-lingual and multi-domain, thus opening up the possibility to assess the true generality of current cross-lingual transfer approaches.   

\subsection{Dialogue State Tracking (DST)}
\label{subsec:dst}

As previously discussed in \S\ref{sec:modular}, DST has recently lost much of its significance for modular \tod due to the ability of Transformer-based models to capture long-distance dependencies and represent the entire dialogue history. For completeness, we briefly summarise the existing multilingual datasets and cross-lingual transfer methods for DST, which are predominantly based on cross-lingual word embeddings. 

\vspace{2mm}
\noindent \textbf{Cross-Lingual Transfer Methods.} The Neural Belief Tracker \shortcite<NBT;>{mrkvsic2017neural,Mrksic:2018acl} is a DST model that updates its internal representation of the state of a conversation at every dialogue turn in a fully data-driven fashion. As such,
NBT was the first neural approach based on word embeddings performing on a par with the models exploiting hand-crafted lexical rules. With the introduction of the multilingual WoZ dataset, \shortciteA{mrkvsic2017multilingualwoz} coupled NBT with cross-lingual word embeddings to enable zero-shot cross-lingual DST transfer. A body of subsequent work on retrofitting CLWEs for pure semantic similarity reported performance gains in cross-lingual transfer for DST, using NBT as the base model \shortcite{vulic2018post,glavavs2018explicit,ponti2018adversarial,ponti2019cross}.  
XL-NBT \shortcite{chen2018xlnbt} resorts instead to multilingual knowledge distillation \shortcite{hinton2015distilling}: the DST knowledge of the English teacher model is transferred to the target language student model by matching their latent representations of parallel words and sentences. A similar technique has been applied more recently for general sentence-level representation learning \shortcite{Reimers:2020emnlp}.

\begin{table}[t]
\resizebox{\textwidth}{!}{
{\footnotesize
\begin{tabular}{llccrc} 
\toprule
\textbf{Dataset}                                               & \textbf{Task}                                                             & \textbf{Languages} & \textbf{Domains}                                                                                                          & \begin{tabular}[c]{@{}c@{}}\textbf{Size}\\\textbf{ (\# dialogues)}\end{tabular} & \textbf{Type}                                          \\ 
\midrule
\rowcolor{Gray}
\begin{tabular}[c]{@{}l@{}}DSTC 5\\ \shortcite{kim2016dstc5} \end{tabular}               & \begin{tabular}[c]{@{}l@{}}dialogue state \\ tracking\end{tabular}        & en, zh               & \begin{tabular}[c]{@{}c@{}}tourist \\ information\end{tabular}                                                           & \begin{tabular}[c]{@{}r@{}}35 [en]\\ 12 [zh]\end{tabular}                       & H2H                                                         \\ 
\hline
\begin{tabular}[c]{@{}l@{}}Multilingual WOZ 2.0\\ \shortcite{Mrksic:2018acl} \end{tabular} & \begin{tabular}[c]{@{}l@{}}dialogue state \\ tracking\end{tabular}        & en, de, it           & \begin{tabular}[c]{@{}c@{}}restaurant \\ booking\end{tabular}                                                            & 1200                                                                            & \begin{tabular}[c]{@{}c@{}}H2H\\ (translated)\end{tabular}  \\ 
\hline
\begin{tabular}[c]{@{}l@{}}DSTC 6\\ \shortcite{hori2019dstc6} \end{tabular}               & \begin{tabular}[c]{@{}l@{}}dialogue \\breakdown \\ detection\end{tabular} & en, ja               & \begin{tabular}[c]{@{}c@{}}chit chat\end{tabular} & \begin{tabular}[c]{@{}c@{}}615 [en]\\ 1696 [ja]\end{tabular}                    & H2M                                                         \\ 
\hline
\begin{tabular}[c]{@{}l@{}}DSTC 9\\ \shortcite{gunasekara2020dstc9} \end{tabular}               & \begin{tabular}[c]{@{}l@{}}dialogue \\state tracking\end{tabular}         & en, zh               & \begin{tabular}[c]{@{}c@{}}7 domains in [en] \\ 5 domains in [zh]\end{tabular}                                           & \begin{tabular}[c]{@{}c@{}}10438 [en]\\ 6012 [zh]\end{tabular}                  & H2M                                                         \\ \hline
\rowcolor{Gray}
\begin{tabular}[c]{@{}l@{}}GlobalWoZ\\ \shortcite{ding2021globalwoz} \end{tabular}               & \begin{tabular}[c]{@{}l@{}}dialogue \\state tracking\end{tabular}         & es, id, zh            &  \begin{tabular}[c]{@{}c@{}}8 domains \\(e.g., restaurants, taxi)\end{tabular} & 1500                 & H2H           \\ 
AllWOZ \shortcite{zuo2021allwoz} & \begin{tabular}[c]{@{}l@{}}dialogue state\\ tracking\end{tabular} & \begin{tabular}[c]{@{}c@{}}fr, vi, pt, \\ko, zh, hi, \\ th\end{tabular}  & \begin{tabular}[c]{@{}c@{}}5 domains \\ (e.g., attraction, hotel, \\ restaurant)\end{tabular} & 700 & \begin{tabular}[c]{@{}c@{}}H2H\\ (translated)\end{tabular} \\ \hline
\rowcolor{Gray} Multi\textsuperscript{2}WOZ \shortcite{hung2022multi2woz} & \begin{tabular}[c]{@{}l@{}}dialogue state\\ tracking\end{tabular} & \begin{tabular}[c]{@{}c@{}}ar, de, ru, zh\end{tabular}  & \begin{tabular}[c]{@{}c@{}}5 domains \\ (e.g., attraction, hotel, \\ restaurant)\end{tabular} & 8,000 & \begin{tabular}[c]{@{}c@{}}H2H\\ (translated)\end{tabular} \\
\bottomrule
\end{tabular}
}%
}
\caption{Multilingual DST datasets. Shaded rows correspond to datasets created from scratch, white rows to datasets translated from English. Abbreviations: human-to-machine (H2M) and human-to-human (H2H). A non-exhaustive list of English DST datasets is given in Table~\ref{table: english  dst} in the Appendix.}
\label{table: mult dst}
\end{table}

The leaderboard of the recent \textit{DSTC 9} challenge \shortcite{gunasekara2020dstc9}, however, indicates that machine translation of the training data coupled with state-of-the-art monolingual DST models in the target language \shortcite{shan2020contextual,kim2020efficient}, known as `translate train', outperforms zero-shot and few-shot cross-lingual transfer based on multilingual representations. On the other hand, DSTC 9 includes only English and Chinese, both endowed with abundant monolingual and parallel data. Hence, this finding may not hold true for other languages. 

\vspace{2mm}
\noindent \textbf{Available Datasets.} \shortciteA{Mrksic:2018acl} translated the WoZ 2.0 DST dataset \shortcite{wen2017network} to German and Italian. Within the dedicated Dialogue State Tracking Challenge (later renamed Dialogue System Technology Challenges), only 3 out of 9 editions to date included multilingual DST tracks. DSTC 5 \shortcite{kim2016dstc5} evaluated DST models on zero-shot cross-lingual transfer from English (training data) to Chinese (development and test data) in the tourism domain. DSTC 6 \shortcite{hori2019dstc6} included a track on dialogue breakdown detection in chat-oriented dialogues, which required transferring knowledge from English to Japanese. 
Finally, as the first challenge to benchmark cross-lingual DST systems on large scale datasets, DSTC 9 \shortcite{gunasekara2020dstc9} included a track focused on transfer between English and Chinese (in both directions), using MultiWOZ 2.1 \shortcite{eric2019multiwoz21} as the English dataset and CrossWOZ \shortcite{zhu2020crosswoz} as the Chinese dataset. Table~\ref{table: mult dst} summarises the key properties of the above-mentioned multilingual DST datasets.

\setlength{\tabcolsep}{4.2pt}
\begin{table}[!t]
\centering
{\scriptsize
\begin{tabular}{llccr} 
\toprule
\textbf{Dataset} & \textbf{Task} & \textbf{Language(s)} & \textbf{Domains}                                                                                     & \begin{tabular}[c]{@{}c@{}}\textbf{Size}\\\textbf{ (dialogues)}\end{tabular} \\ 
\midrule
\rowcolor{Gray}
\begin{tabular}[c]{@{}l@{}}CrossWOZ \shortcite{zhu2020crosswoz}\end{tabular}       & \begin{tabular}[c]{@{}l@{}}dialogue state\\ tracking;\\ end-to-end;\end{tabular} & zh                                                                       & \begin{tabular}[c]{@{}c@{}}5 domains\\ (e.g., attraction, \\ hotel, taxi)\end{tabular}      & 6012  \\ 
\hline
\rowcolor{Gray}
\begin{tabular}[c]{@{}l@{}}RiSAWOZ \shortcite{quan2020risawoz}\end{tabular}        & \begin{tabular}[c]{@{}l@{}}dialogue state\\ tracking;\\ end-to-end;\end{tabular} & zh                                                                       & \begin{tabular}[c]{@{}c@{}}12 domains\\ (e.g., education, \\ car, hospitality)\end{tabular} & 11200 \\ 
\hline
WMT 2020 Chat \shortcite{farajian-etal-2020-findings} & end-to-end & de & \begin{tabular}[c]{@{}c@{}}6 domains \\ (e.g., ordering pizza,\\ movie tickets)\end{tabular} & 692 \\ \hline
\rowcolor{Gray}
ViWOZ \shortcite{van2022viwoz} & \begin{tabular}[c]{@{}l@{}}NLU; dialogue\\ state tracking;\\ end-to-end;\end{tabular} & vi & \begin{tabular}[c]{@{}c@{}}7 domains \\ (e.g., attraction, bas,\\ hospital, hotel)\end{tabular} & 5000 \\ \hline
BiTOD \shortcite{lin2021bitod} & end-to-end & en, zh & \begin{tabular}[c]{@{}c@{}}5 domains \\ (e.g., attractions, hotel, \\ restaurant, weather)\end{tabular} & 7,232 \\ \hline
COD \shortcite{majewska2022cross} & \begin{tabular}[c]{@{}l@{}}NLU; dialogue\\ state tracking;\\ end-to-end;\end{tabular} & \begin{tabular}[c]{@{}c@{}}en, ar, \\id, ru, sw\end{tabular}  & \begin{tabular}[c]{@{}c@{}}11 domains \\ (e.g., flights, homes, \\ music, payments)\end{tabular} & 368 \\
\bottomrule
\end{tabular}
}%
\caption{Multilingual datasets for end-to-end training. The table also includes monolingual non-English datasets. Shaded rows correspond to datasets created from scratch, white rows to datasets translated from English. A non-exhaustive list of English datasets is provided in the Appendix.}
\label{table: multi e2e}
\end{table} 
\setlength{\tabcolsep}{6pt}

\subsection{Natural Language Generation (NLG)}
\label{subsec:nlg}

 In contrast to other sub-tasks of modular \tod, multilingual response generation for \tod has received limited attention. We thus take a broader look at multilingual NLG in general. 

\vspace{2mm}
\noindent \textbf{Traditional NLG.} Traditionally, the production of sentences in languages other than English, conditioned on structured or continuous representations of the desired meaning, relied on NLG pipelines \shortcite{reiter1997building,reiter2000building}. The last module of such pipelines is a linguistic (surface) realiser, responsible for outputting the final surface text based on language-specific morpho-syntactic and orthographic constraints (e.g., word order, inflectional morphology). These constraints can be enforced through hand-crafted grammar-based systems \shortcite{gatt2018survey,bateman1997enabling,elhadad1996overview}, manually created templates \shortcite{mcroy2003augmented}, and statistical methods \shortcite{filippova-strube-2007-generating}.
To facilitate the general usage of grammar-based systems, characterised by a high level of linguistic detail, simpler realisation engines that provide syntax and morphology APIs have been developed \shortcite{gatt2009simplenlg} and subsequently adapted to a number of languages, including Spanish \shortcite{ramos2017adapting}, Galician \shortcite{cascallar-fuentes-etal-2018-adapting}, Italian \shortcite{mazzei2016simplenlg}, German \shortcite{bollmann2011adapting}, Brazilian Portuguese \shortcite{de2014adapting}, and French \shortcite{vaudry2013adapting}. A hybrid approach coupling linguistic knowledge (i.e., a grammar and a lexicon) with statistical methods was recently proposed by \shortciteA{GARCIAMENDEZ2019372}.

\vspace{2mm}
\noindent \textbf{Translation-Based Cross-lingual Transfer.} 
Given the reliance of data-driven NLG models on the availability of training data and their scarcity in the vast majority of the world's languages, cross-lingual transfer methods have been leveraged to enable NLG in low-resource scenarios. They have been employed in two main ways (cf.\ \S\ref{sec:gaps}): i) in a `translate test' setting, an NLG system is trained on English. At evaluation time, the input text in a target language is translated into English before feeding it to the system, and the English text generated in output is translated back to the target language \shortcite{wan2010cross}. Alternatively, ii) in a `translate train' setting, the English training data is translated into the target language, which provides supervision to learn an NLG system directly in the target language \shortcite{shen2018zero,duan2019zero}. 

\vspace{2mm}
\noindent \textbf{Cross-lingual Transfer via Multilingual Representations.} Pretraining of multilingual sequence-to-sequence models, initially devised for neural machine translation \shortcite{liu2020multilingual,lin2020pre,kim2020and}, has been successfully applied to facilitate cross-lingual transfer in other NLG applications as well. For example, \shortciteA{kumar-etal-2019-cross} first pretrain a bilingual English-Hindi model through unsupervised MT (via denoising autoencoding and back-translation); they then fine-tune the model for question generation on both (large) English and (small) Hindi data. 
Similarly, \shortciteA{chi2020cross} pretrain a Transformer-based encoder-decoder on several languages, with denoising auto-encoding and cross-lingual masked language modelling objectives, and fine-tune it for question generation and abstractive summarisation. 
%
However, recent work also exposed a series of limitations of massively multilingual Transformer-based encoders.
Adopting the approach of \shortciteA{wang-cho-2019-bert}, \shortciteA{ronnqvist2019multilingual} evaluated mBERT \shortcite{devlin2019bert} on NLG tasks in English, German, Danish, Finnish, Norwegian (Bokmal and Nynorsk) and Swedish. They found that massively multilingual models (1) significantly under-perform their monolingual counterparts and (2) struggle in handling complex morphology. This somehow undermines the viability of MEs for large-scale multilingual NLG, although more recent work suggests that the general task performance of the MEs can be partially recovered through language-specific adaptation \shortcite{Rust:2020arxiv}.


\vspace{2mm}
\noindent \textbf{Available Datasets.} Training data for NLG in languages other than English remain scarce: small datasets are available in Korean \shortcite{chen2010training}, Spanish \shortcite{GARCIAMENDEZ2019372}, and Czech \shortcite{duvsek2019neural}. There exist also structured data-to-text datasets for German and French \shortcite{nema-etal-2018-generating} and image-to-description datasets in Chinese \shortcite{li2016adding} and Dutch \shortcite{miltenburg2017cross,miltenburg2018DIDEC}, as well as cross-lingual English-German data \shortcite{elliott-EtAl:2016:VL16}.

\subsection{End-to-End Dialogue}
\label{subsec:e2e}

Lately, end-to-end dialogue modelling has emerged as a promising alternative to modular \tod. Most of the algorithms fall within the sequence-to-sequence (seq2seq) framework, reading user utterances in input and generating system responses in output \shortcite{wen2017network,madotto2018end2endknowledge,ham2020end2endgpt}.
Unfortunately, training reliable seq2seq models demands extensive supervision, making end-to-end dialogue systems data-hungry. However, collecting task-oriented dialogues is much more expensive and laborious than collecting open-domain conversations for training chatbots. 
As a result, only a few monolingual end-to-end \tod system have been developed in languages other than English, such as Chinese \shortcite{zhu2020crosswoz,quan2020risawoz}. Future work should also look into different flavours of E2E modelling in multilingual context, e.g., distinguishing between (i) \textit{strict E2E} models, where the system leverages only natural dialogue texts, and (ii) \textit{weak E2E} models, where the system is additionally allowed to rely on intermediate dialogue annotations (e.g., dialogue states and dialogue acts).

Finally, we list the available datasets for non-English e2e \tod in Table \ref{table: multi e2e}.
Although this survey concentrates on \tod, we additionally list available datasets for multilingual open-domain dialogue in the Appendix. 

%% file: 4-challenges.tex
\section{Open Challenges and Future Directions}
\label{sec: future directions}
In light of the current state of multilingual \tod research, covered in \S\ref{sec:stquo}, we now identify the main open challenges, hint to some of their possible solutions, and conjecture future directions of research. In particular, in \S\ref{ssec:lingdiv} we analyse the linguistic diversity, idiomacity, and cultural adaptation of current datasets for multilingual \tod, the aspects which, we argue, should be prioritised in future resource creation endeavours. In \S\ref{ss:lowres}, we argue that cross-lingual transfer in \tod NLU can adopt solutions devised for related NLP tasks, especially to deal with low-resource scenarios. Furthermore, we consider the compounded difficulties that multilinguality adds for fluency in text generation (especially with respect to complex morphology and code switching) in \S\ref{ssec:fluencygen} and for collecting human evaluation in \S\ref{ss:eval}. Finally, in \S\ref{ss:speech} we broaden the scope of our survey to multilingual speech for voice-based \tod.

\subsection{Outlook for Multilingual \tod Datasets}
\label{ssec:lingdiv}

\noindent \textbf{Linguistic Diversity.}
Evaluating on representative language samples is key for long-term development of multilingual \tod systems. In fact, 
the purpose of multilingual datasets is to assess the expected performance of a model across languages \shortcite{hu2020xtreme,Liang:2020xglue}. If all the languages in the evaluation sample are similar, cross-lingual transfer is simplified and the resulting estimates could be overly optimistic  \shortcite{ponti2020xcopa}. 
Instead, 
the sample of languages should ideally be diverse in terms of language family, area, and typological features \shortcite{ponti2019modeling}.

\begin{table}[t]
\centering
\def\arraystretch{0.91}
\resizebox{\textwidth}{!}{
{\footnotesize
\begin{tabular}{l ccccc ccc}
\toprule
             & \multicolumn{1}{l}{M. TOP} & \multicolumn{1}{l}{M. ATIS} & \multicolumn{1}{l}{MultiATIS++} & \multicolumn{1}{l}{MTOP} & \multicolumn{1}{l}{xSID} & \multicolumn{1}{l}{XCOPA} & \multicolumn{1}{l}{TyDi QA} & \multicolumn{1}{l}{XNLI} \\ \cmidrule(lr){2-6} \cmidrule(lr){7-9}
\# languages & 3                                     & 3                                      & 9                                & 6  &       13                & 11                         & 11                          & 15                        \\ \cmidrule(lr){2-6} \cmidrule(lr){7-9}
Typology     & 0.20                                 & 0.29                                  & 0.33                            & 0.29 &      0.37              & 0.41                       & 0.41                        & 0.39                      \\ 
Family       & 0.67                                 & 0.67                                  & 0.44                            & 0.33 &          0.50          & 1.0                        & 0.9                         & 0.5                       \\ 
Macroareas       & 0                                 & 0                                 & 0                            & 0          &   0        & 1.67                        & 0.92                       & 0.37                      \\ \bottomrule
\end{tabular}
}%
}
\caption{Diversity indices of multilingual dialogue NLU datasets in terms of typology, family, and macroarea. Linguistically diverse datasets of several other NLP tasks shown for comparison: commonsense reasoning (XCOPA; \shortciteR{ponti2020xcopa}), natural language inference (XNLI; \shortciteR{conneau2018xnli}) and QA (TyDI QA; \shortciteR{clark2020tydi}). For the description of the three diversity measures, we refer the reader to \shortciteA{ponti2020xcopa}.}
\label{table: linguistic diversity nlu}
\end{table}

NLU is the only component of modular \tod whose datasets cover an extensive sample of languages. We quantify the linguistic diversity of these datasets through the metrics proposed by \shortciteA{ponti2020xcopa}, which are based on typological, family and geographical properties.\footnote{To measure typological diversity, we calculate the entropy of 103 binary typological features from URIEL \shortcite{littell2017uriel} across languages, and then average across features. 
To measure family diversity, the number of distinct families represented in the dataset is divided by the total number of languages in the dataset. To measure geographical diversity, we calculate the entropy of the distribution across geographical macroareas of the dataset languages.} The sample diversity scores are shown in Table~\ref{table: linguistic diversity nlu}. For comparison, we include the most diverse datasets for other NLP tasks, such as natural language inference (XNLI; \shortciteR{conneau2018xnli}), question answering (TyDi QA; \shortciteR{clark2020tydi}), and causal commonsense reasoning (XCOPA; \shortciteR{ponti2020xcopa}). 

From Table~\ref{table: linguistic diversity nlu}, several shortcomings of the language samples within existing multilingual dialogue NLU datasets become apparent. First, they mostly originate from a single macroarea, namely Eurasia. The only exception is xSID, which contains Indonesian from Papunesia. Second, they fall short of representing a significant variety of linguistic phenomena. In particular, the languages are identical with respect to 23 out of 103 typological URIEL features. Third, only MultiATIS++ and xSID cover a number of languages comparable to non-dialogue NLP datasets, but even so, the majority of them belong to the same family (Indo-European). Datasets for other NLP tasks can therefore serve as a polestar for the selection of more diverse language samples in dialogue NLU.

Worse yet, beyond NLU, there is currently a complete lack of large dialogue datasets with full-fledged multi-turn conversations in multiple typologically diverse languages (see Table~\ref{table: multi e2e}). If such a dataset existed, it would enable end-to-end training of multi-turn \tod systems in multiple languages and widen our understanding of similarities and differences across languages at the level of entire dialogues.
In conclusion, the collection of (possibly end-to-end) datasets covering a \textit{much wider} set of families, macroareas and typological properties could define the path to future milestones in multilingual \tod.

\vspace{2mm}
\noindent \textbf{Idiomacity and Cultural Adaptation.} 
In addition to linguistic diversity, another limitation of current multilingual \tod datasets stems from their creation process. In fact, training and evaluation instances are often directly translated from English. However, this might cause unwanted ramifications. First, `translationese', the language variety of translated texts, is different from that of natural and spontaneous texts. This is due to both universal patterns in translation (simplification, normalisation, and explicitation) and linguistic interference: the source language spills its lexical and structural properties over the target language \shortcite{lembersky2012language,volansky2015features}. These artefacts, introduced by the translation procedure, could make the dataset not representative of real-life dialogue and cultural context of the target language \cite{hershcovich2022challenges} and instead give an edge to translation-based cross-lingual transfer. Hence, the evaluation performance becomes unreliable and excessively optimistic \shortcite{artetxe2020translation}. \shortciteA{koppel-ordan2011:acl} studied the differences between translated-into-English and original English texts. They demonstrate that there is a significant difference in lexical characteristics of the texts: e.g., there are some stark differences in the frequency of usage of functional words and pronouns. Recent work by \shortciteA{majewska2022cross} presents a qualitative analysis in the context of dataset creation for multilingual \tod, comparing dialogue data obtained via translation and free-form generation by native speakers of the target language. The paper presents multiple examples of the bias from English on both lexical and structural syntactic level. We refer the reader to the paper for some concrete examples.

Secondly, the information and the topics touched upon in a conversation (e.g., in the domain of airline travels, names of destinations or flight companies) may vary across cultures and locales. However, translation-based approaches reflect the perspective of the English-speaking culture, its `presupposed' factual knowledge, and the worldview of its community of speakers \shortcite{clark2020tydi}. For these reasons, \tod benchmarks should be ideally based on original utterances grounded in the appropriate locale and culture. 

Recognising this need, there have been some very recent developments in the direction of creating localised and culturally adapted \tod datasets. Namely, \shortciteA{ding2021globalwoz} undertake an automatic approach to localisation in which the English slots are substituted by their local counterparts, obtained via Web-crawling suitable values. In contrast, \shortciteA{majewska2022cross} ask the native speakers to generate utterances based on templatic dialogue outline. The dialogue creators are encouraged to substitute English slot values with their target-language counterparts. While the former approach presents a more controlled automatic localisation procedure,  the latter is more human-driven, providing a closer contact with the local community speaking the target language. Furthermore, \shortciteA{fitzgerald2022massive} localise their multilingual NLU dataset in two stages. In the first stage, native speakers were asked to translate and localise slot values; in the second stage, another group of human subjects translated or localised the entire phrase using the
slot task output provided by the first worker. Going beyond research in \tod, \shortciteA{hershcovich2022challenges} suggest that collecting multilingual data within (large) local communities results in culturally richer data and avoids imposing English-driven use cases. Additionally, it can reduce the per-person manual effort (by dividing the work between more people) which is often a bottleneck in the data collection process. Other potential approaches to reducing manual data creation and curation effort are briefly discussed below.

\vspace{2mm}
\noindent \textbf{Reducing Human Effort in \tod Data Curation.} Collecting dialogue datasets is notoriously hard, time-consuming and requires a lot of manual labour. In multilingual \tod, the problem scales proportionally with the number of languages. It is thus natural to seek for methods to reduce manual human effort without compromising data quality.

Firstly, several automated approaches to reduce human effort were proposed, predominantly focusing on data augmentation. These approaches, previously applied to multilingual dialogue, include (i) word or span substitution, creating code-switched data between source and target languages \shortcite{liu2019attention,krishnan2021multilingual,qin2021cosda} or synonymous span substitution in a target language \cite{louvan2020simple}2020; and (ii) semi-supervised training with target language sentence retrieval \shortcite{razumovskaia2022:acl}. 

Secondly, since direct translation still dominates multilingual \tod data collection, there have been several approaches to lower human effort in the translation procedure. In most cases translators would simultaneously annotate the datasets with slot labels and/or dialogue states, depending on the tasks the dataset covers \shortcite{mrkvsic2017multilingualwoz,xu2020multiatis++,van-der-goot-etal-2021-masked}. 
One approach simplifies the translation process itself, which typically proceeds in two stages: (i) machine translation into the target language; (ii) manual post-editing by native speakers of the language \cite{zuo2021allwoz,hung2022multi2woz}. One can also reduce the amount of manual annotation work by projecting labels from the source language to the target language. While label projection for sentence classification tasks is trivial, it is harder but still attainable for the word- and context-level tasks. Some of the existing methods are unsupervised and rely on parallel corpora \cite{dyer2013simple,dou2021word}. Other methods complement machine translation with a word alignment step \cite{Jain:2019emnlp}. Once the labels have been automatically projected, one could hire annotators, native speakers of the target language, only for editing the projected slots.








\subsection{Coping with Low-Resource Scenarios in NLU}
\label{ss:lowres}

\noindent \textbf{Parallels with Other NLP Tasks.}
As discussed in \S\ref{sec:stquo}, intent detection is a standard classification task, which can also be recast as a question answering task \shortcite{Namazifar:2020qanlu}. On the other hand, slot filling can be framed as a sequence labelling \shortcite{Louvan:2020coling} or a span extraction task \shortcite{coope2020restaurants8k,Henderson:2021naacl}. Along the same lines, DST is sometimes formulated as a semantic parsing task in monolingual multi-domain settings \shortcite{cheng2020dstassemparsing}: dialogue states are represented as a hierarchical semantic structure which includes information about the domain, past actions, and the slots filled or requested.\footnote{Formulating DST as semantic parsing opens up several paths for future research. First, structured representations naturally allow for semantic compositionality and cross-domain knowledge sharing. In multilingual \tod, they could similarly allow for cross-lingual knowledge sharing. Secondly, structured representations facilitate the integration of external knowledge. For instance, tables like those widely adopted in semantic parsing \shortcite{zhu2020semanticparsingtables,sun2019semanticparsingwebtables} may improve cross-lingual slot labelling by mapping expressions for named entities (e.g., city names) across languages.}

This effectively means that the standard methodological `machinery' forged to handle low-resource languages and domains can be directly applied to joint multilingual modelling and cross-lingual transfer in \tod NLU \shortcite{ponti2019modeling,Hedderich:2021naacl}. In what follows, we provide a very brief and non-exhaustive overview of promising cutting-edge techniques that might also benefit low-resource \tod.\footnote{For a comprehensive survey of methods for low-resource NLP, we refer the reader to \shortciteA{Hedderich:2021naacl}.} The reader should, however, still bear in mind the core deficiencies of the current methodology in relation to multilingual \tod, as previously discussed in \S\ref{sec:gaps}.

Low-resource languages should profit from annotated resources in higher-resource languages. Besides translation-based transfer \shortcite{upadhyay2018multiatis,schuster2019cross,hu2020xtreme}, annotations can be propagated source-to-target using parallel data and word alignments \shortcite{Ni:2017acl,Jain:2019emnlp,xu2020multiatis++}. Annotation and model transfer can also be realised via cross-lingual word embeddings \shortcite{glavavs2019properly,Ruder:2019jair}. 
Recently, unmatched performance in cross-lingual transfer has been achieved by multilingual Transformer-based encoders such as multilingual BERT \shortcite{devlin2019bert}, XLM-R \shortcite{Conneau:2020acl}, and mT5 \shortcite{mt5}, or cross-lingual alignment of their monolingual counterparts \shortcite{Schuster:2019naacl,Conneau:2020emerging,Cao:2020iclr}. These models can be additionally (i) extended to cover radically low-resource languages \shortcite{Pfeiffer:2020emnlp,ponti2020xcopa,Hedderich:2020emnlp,Ustun:2020emnlp} and even languages with unseen scripts \shortcite{Pfeiffer:2020unks}, or (ii) refined via few-shot learning on a small subset of target-language annotated examples \shortcite{lauscher2020zero,Bhattacharje:2020emnlp}. However, the current performance of cross-lingual transfer to low-resource target languages (e.g., African languages, indigenous languages of the Americas) still lags dramatically behind that of transfer to high-resource target languages \shortcite{lauscher2020zero,Wu:2020repl,Zhao:2020arxiv}.

Pretrained language models can also be adaptively fine-tuned with unannotated in-domain data in both the source and the target language to pick up more task-specific knowledge, which typically results in slight performance gains \shortcite{henderson2019convert,Gururangan:2020acl}. Along the same lines, the entire research area focusing on domain adaptation in NLP \shortcite{Kim:2018acl,Ziser:2018emnlp,Ruckle:2020emnlp,Ramponi:2020coling} can also offer direct guidance on how to leverage high-resource \tod domains to boost performance in resource-lean \tod domains. Note that in multilingual \tod, we typically encounter formidably difficult ``double-scarce'' setups, simultaneously dealing with both low-resource domains \textit{and} low-resource languages.

Another plausible solution to data paucity is making the multilingual \tod NLU models more robust in low-resource scenarios through data augmentation \shortcite{Du:2020arxiv,Xie:2020nips}: (i) at the token level with synonymy-based substitutions generated automatically \shortcite{Kobayashi:2018naacl,Gao:2019acl} or taken from lexico-semantic resources \shortcite{Raiman:2017emnlp,Zou:2019emnlp,Dai:2020coling}, or rule-based morphological inflection \shortcite{Vulic:2017acl,Vania:2019emnlp}, (ii) at the sentence level by manipulating dependency trees \shortcite{Ponti:2018acl,Sahin:2018emnlp}, back-translating \shortcite{Edunov:2018emnlp}, or generating synthetic adversarial examples \shortcite{Garg:2020emnlp,Morris:2020emnlp}; (iii) at the annotation level by automatically labelling more sentences, filtering them, and using them as silver training data \shortcite{Onoe:2019naacl,Du:2020arxiv}. 

\begin{figure*}[t!]%
    \centering
    \subfloat[\centering Intent classification] {{\includegraphics[width=0.45\textwidth]{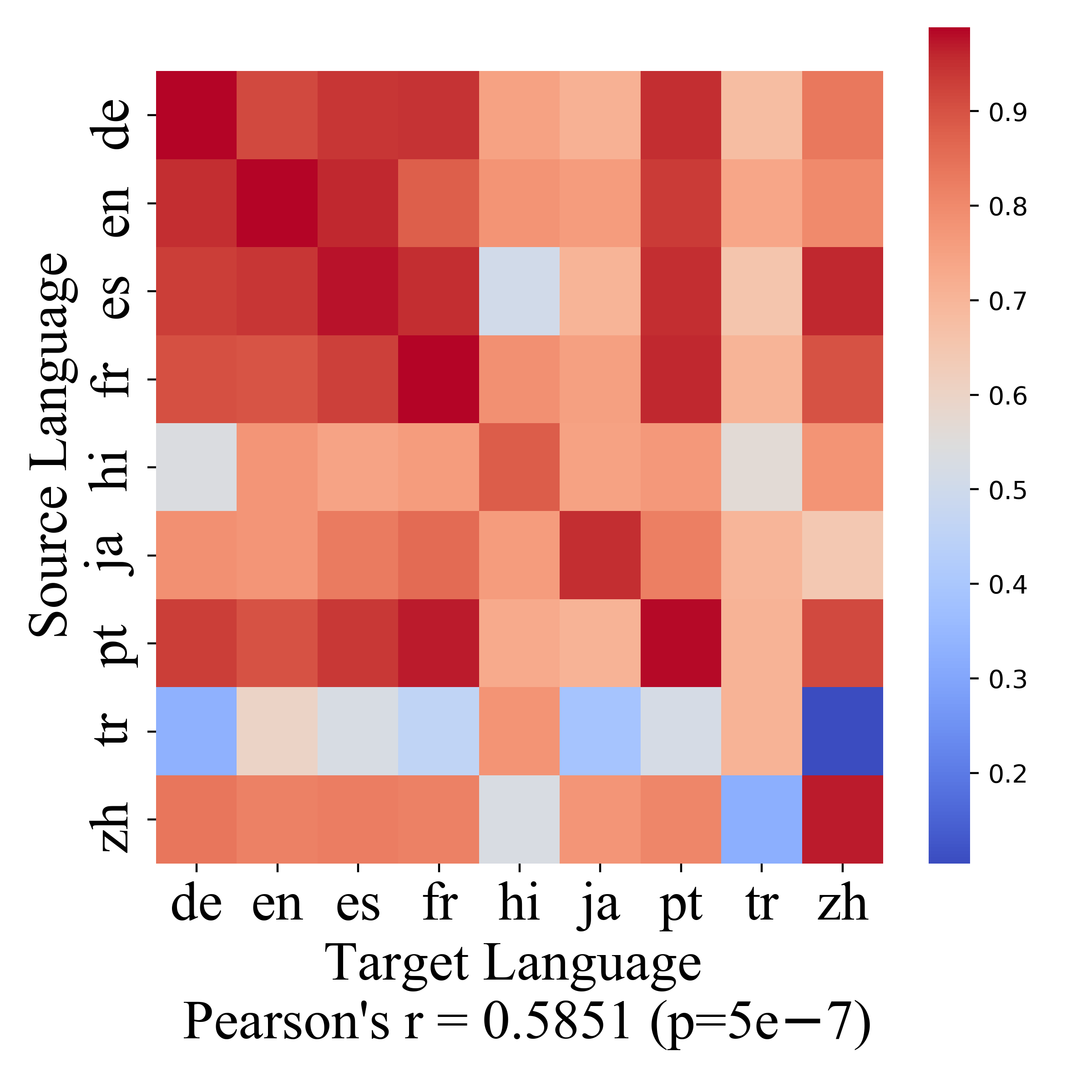}}}%
    \qquad
    \subfloat[\centering Slot labelling ]{{\includegraphics[width=0.45\textwidth]{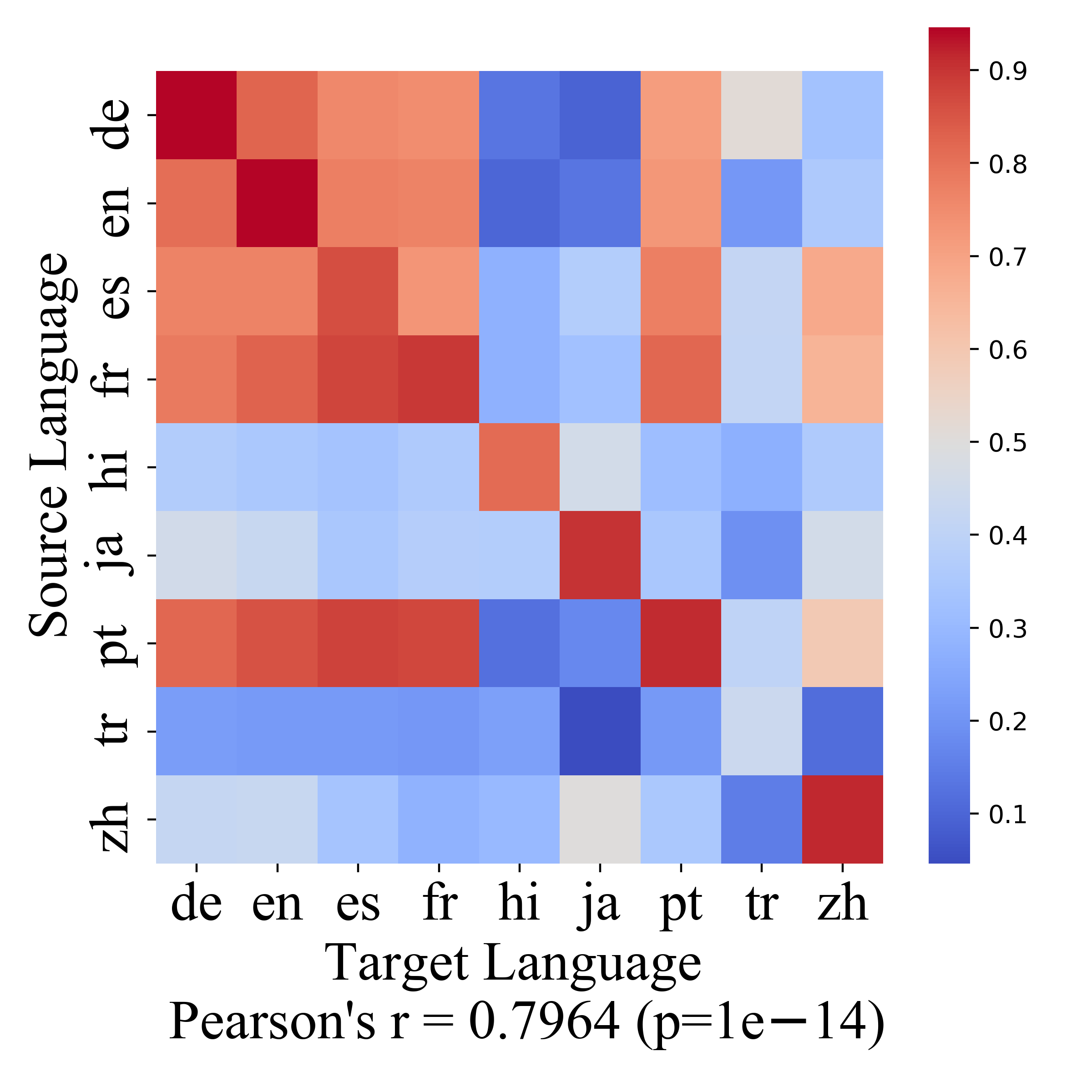} }}%
    \vspace{-1.5mm}
    \caption{Pair-wise cross-lingual transfer performances on intent classification (left) and slot labelling (right) between each source language (y axis) and each target language (x axis).}%
    \label{fig:ling sim source target}%
\end{figure*}
\begin{table}[!t]
\centering
\resizebox{0.65\textwidth}{!}{%
\begin{tabular}{l clcl} 
\toprule
\textbf{Feature Group} & \multicolumn{2}{c}{\textbf{Intent Classification}} & \multicolumn{2}{c}{\textbf{Slot Labelling}}        \\ 
\toprule
                       & Pearson's $r$ & p-value $\leq$          & Pearson's $r$ & p-value $\leq$          \\\cmidrule(r){2-3} \cmidrule(r){4-5}
Phonology              & $-0.2467$                       & \num{5e-2} 
& $-0.2504$                       
& \num{5e-2} 
\\
Geography              & $-0.2263$                       & \num{5e-1} 
& $-0.3270$                       & 
\num{5e-2} 
\\
Phylogenetics          & $-0.4895$                       & \num{5e-4}     & $-0.6122$                       & \num{5e-7}   \\
Syntax                 & $-0.5131$                       & \num{5e-5}   & $-0.6919$                       & \num{5e-11}  \\
\bottomrule
\end{tabular}
}
\caption{Correlation between single-source zero-shot transfer performance in NLU and source--target language distance according to different groups of properties. Phonology and syntax features come from URIEL \shortcite{littell2017uriel}, phylogenetic features from Glottolog \shortcite{hammarstrom2017glottolog}.}
\label{table: ablation study linguistic features}
\end{table}

Distant and weak supervision methods \shortcite{Luo:2019acl,Alt:2019acl}, often leveraged to share knowledge between structurally similar low-resource NER models \shortcite<e.g.,>{Cao:2019emnlp,Mayhew:2019conll,Lison:2020acl}, might also prove beneficial for slot filling, as a sequence labelling task. 
Meta-learning frameworks such as MAML \shortcite{Finn:2017icml} have also emerged recently as strategies to tackle low-resource cross-lingual and cross-domain transfer for several NLP classification tasks \shortcite{Farhad:2020emnlp,Heijden:2021eacl}; however, they are yet to find their application in (multilingual) \tod systems. 


While taking the inspiration from related NLP tasks is a natural approach to speed up the progress in multilingual \tod, it is important to also take into account the aspects where the development of \tod systems differs from working with related NLP tasks. Firstly, the models need to operate with and integrate (multi-speaker) dialogue history as conversational context, which implies processing long sequences.  Most pretrained Transformer-based models cannot handle such long input, as they are limited by the self-attention operation which scales quadratically with the length of the input sequence. Models such as LongFormer \shortcite{beltagy2020longformer} and BigBIRD \shortcite{zaheer2020big} propose modifying the self-attention with localised attention and sparse attention, respectively, for long sequence processing. These efficient Transformer-based models are yet to find their application in the context of multilingual \tod. Further, such heuristic solutions might still be suboptimal for languages with word order and manner of speaking which is radically different to English and other major languages \cite{Ruder:2022acl}. 

The inherent conversational nature of \tod tasks poses additional challenges to direct transfer of methods developed for other NLP tasks to multilingual dialogue. For instance, models pretrained with response selection or retrieval tasks on conversational data are more suited for modelling dialogue than their counterparts pretrained with language modelling task \shortcite{mehri2020dialoglue,coope2020restaurants8k}. This indicates that we might witness substantial improvements in multilingual dialogue if standard multilingual encoders are replaced with ones pretrained or fine-tuned on conversational data with dialogue-oriented objectives \cite{henderson2019convert,hung2022multi2woz}. There are also linguistic phenomena (e.g., code switching) which are more widespread in the colloquial speech than in the written format. The methods developed for multilingual \tod must also be able to deal with utterances containing such phenomena (see also later \S\ref{ssec:fluencygen}). 

\vspace{2mm}
\noindent \textbf{Source Selection for Cross-lingual Transfer.} 
When porting a dialogue system to new languages, zero-shot transfer is an effective method to bypass costly data collection and annotation for every target language. However, based on prior work in general-purpose cross-lingual NLP, we detect three crucial gaps which require more attention in future work and that may play an instrumental role in the final task performance: \textbf{1}) the choice of source language(s)  \shortcite{zoph2016transfer,dabre2017empirical,lin2019choosing}, as recently hinted at for multilingual \tod NLU by \shortciteA{krishnan2021multilingual}; \textbf{2}) harnessing multiple source languages rather than a single one \shortcite{zoph2016transfer,pan2020multi,wu2020single,ansell-etal-2021-mad-g}; \textbf{3}) few-shot transfer with a small number of target-language examples, as opposed to fully zero-shot transfer \shortcite{lauscher2020zero}.
In other words, the go-to option of always transferring from English in a zero-shot fashion might be sub-optimal for a large number of target languages.

In order to adduce evidence in support of these two conjectures, we conduct a series of preliminary empirical studies. 
For our experiments, we focus on the two core dialogue NLU tasks, intent classification and slot filling, as documented in MultiATIS++ (see Table~\ref{table:nlu multilingual}). As a neural architecture, we adopt the \textit{de facto} standard of a pretrained encoder (multilingual BERT) and a classifier head (see \ref{subsec:nlu}).\footnote{As the baseline model, uncased multilingual version of BERT-base \cite{devlin2019bert} is used. We use Adam optimizer \cite{kingma2015adam} with learning rate 5e-5 and warm-up ratio  of 0.1. The models are trained with the batch size of 32 for 20 epochs. The models are implemented using gluonNLP package \cite{guo2020gluoncv}.} 

\begin{table}[!t]
\centering
\resizebox{\textwidth}{!}{%
\begin{tabular}{l cccccccccc} 
\toprule
                  & de     & en     & es     & fr     & hi     & ja     & pt     & tr     & zh     & AVG     \\ 
\midrule
\rowcolor{Gray}
& \multicolumn{10}{c}{\sc Zero-shot} \\
\textit{typ-closest} & 95.30 & 91.60 & 93.84 & 96.87 & 78.11 & 77.72 & 95.30 & 67.80 & 64.39 & 75.91  \\
\textit{multi-source}      &   97.20     &    93.95    & 97.20 & 89.59 & 84.61 & 86.11 & 92.16 & 80/57 & 82.31 &     89.30    \\
\textit{ensemble}          &    92.50    &    90.26    &    96.64    &    95.18    &    77.88    &    77.04    &    95.30    &    75.04    &    84.99    &    87.20     \\
\rowcolor{Gray}
& \multicolumn{10}{c}{\sc Few-shot} \\
\textit{typ-closest} & 94.66 & 95.07 & 94.92 & 95.19 & 88.86 & 91.08 & 94.29 & 87.61 & 92.35 & 92.67 \\
\textit{multi-source} & 95.78 & 96.38 & 94.92 & 95.41  & 91.56  & 92.98 & 93.95 & 89.13 &  93.43 &  93.73 \\
\bottomrule
\end{tabular}
}
\caption{Intent classification results (accuracy) on MultiATIS++ \shortcite{xu2020multiatis++} for 3 transfer methods. Few-shot results are averaged across 3 runs.}
\label{tab: multi source intent classification}
\end{table}
\begin{table}[!t]
\centering
\resizebox{\textwidth}{!}{%
\begin{tabular}{l cccccccccc} 
\toprule
                  & de     & en     & es     & fr     & hi     & ja     & pt     & tr     & zh     & AVG     \\ 
\midrule
\rowcolor{Gray}
& \multicolumn{10}{c}{\sc Zero-shot} \\
\textit{typ-closest} & 80.87 & 82.50 & 88.20 & 87.27 & 22.96 & 50.11 & 77.48 & 51.11 & 45.80 &  65.14 \\
\textit{multi-source}      &   87.45     &    91.05    & 86.39 & 82.61 & 68.25 & 74.92 & 87.25 & 60.09 & 69.03 &    78.56     \\
\textit{ensemble}          &   82.75     &    85.05    &   77.56     &   76.19     &  14.14      &    9.44    &       74.00 &   45.63     &   37.29     &      55.78   \\
\rowcolor{Gray}
& \multicolumn{10}{c}{\sc Few-shot} \\
\textit{typ-closest} & 91.29 & 92.67 & 84.93 & 86.56 & 81.99 & 86.45 & 88.43 & 73.44 & 86.57 &  85.81 \\
\textit{multi-source} & 92.51 & 93.76 & 86.25 & 87.43  & 85.46 & 87.47  & 89.08 & 81.34 & 88.29 &  87.95 \\
\bottomrule
\end{tabular}
}
\caption{Slot labelling results (F-1) on MultiATIS++ \shortcite{xu2020multiatis++} based on 3 transfer methods. Few-shot results are averaged across 3 runs.}
\label{tab: multi source slot labelling}
\end{table}

First, we investigate, for every target language in MultiATIS++, its most compatible source language for cross-lingual transfer. We fine-tune our model on a single source at a time and evaluate it in a zero-shot setting on all targets. The resulting performances are pictured in the heat maps of Figure~\ref{fig:ling sim source target}. It emerges that English (\textsc{en}) is not always the best source language (e.g., it is surpassed by \textsc{es} for transfer to \textsc{fr} in intent classification). Moreover, the set of suitable sources is task-dependent. In particular, the knowledge needed for intent classification is more amenable to be transferred to languages outside a language family (\textsc{ja, tr, zh} are not Indo-European) and with different scripts (\textsc{hi, ja, zh} are not written with the Latin alphabet) than slot filling, which is more language-sensitive. As a consequence, scores in intent classification have generally a smaller variance. Finally, as expected, the training dataset size plays a significant role: Turkish and Hindi may count on less annotated examples, which partially explains their lower figures. 

To shed further light onto source selection, we use the cosine similarity between URIEL feature vectors \shortcite{littell2017uriel}, which capture syntactic, phonological, geographic and phylogenetic language properties, as a proxy for the similarity between languages. Based on this metric, we measure the Pearson's correlation between linguistic similarity and the cross-lingual transfer performances reported in Figure~\ref{fig:ling sim source target}. 
The results of this analysis are summarised in Table~\ref{table: ablation study linguistic features}. We discover moderate and strong correlations between language similarity and performance in intent classification and slot labelling, respectively. The highest correlation coefficients are found for phylogenetic and syntactic properties, which can be therefore considered reliable indicators of compatibility for language pairs.


Secondly, we run an additional set of experiments on intent classification and slot filling in MultiATIS++, in order to compare alternative cross-lingual transfer methods and data paucity regimes.
In particular, for every target language we compare three transfer methods: i) \textit{typ-closest}, where we select the model fine-tuned on the typologically closest source language, as determined by the cosine similarity of URIEL \shortcite{littell2017uriel} typological features;\footnote{A table of the typologically closest language for each target language is provided in the Appendix.} ii) \textit{multi-source}, where the model is fine-tuned on all languages but the target; iii) \textit{ensemble}, where the predictions are determined by 8 independent models, each separately trained on a single source language, through majority voting. As for different data paucity regimes, in addition to zero-shot transfer, we also consider few-shot adaptation in the spirit of \shortciteA{lauscher2020zero}: for every target language, we continue fine-tuning its \textit{typ-closest} or \textit{multi-source} models on 500 additional examples sampled from its training split.

The results for intent prediction and slot labelling are summarised in Tables~\ref{tab: multi source intent classification}--\ref{tab: multi source slot labelling}. Comparing different transfer methods, \textit{multi-source} has a significant edge over both \textit{typ-closest} (+13.39 points in accuracy for intent classification, +13.42 in $F_1$ score for slot filling) and \textit{ensemble} (+2.10 for intent classification, +22.78 for slot filling). Note that the performance drop of \textit{ensemble} model in slot labelling proves again its sensitivity to the source languages, as the majority of the voters is unreliable. However, jointly training over all sources not only remedies to this, but further cements the cross-lingual synergy by letting languages borrow statistical strength from one another. This corroborates previous findings \shortcite{xu2020multiatis++} that multi-source fine-tuning is beneficial even when full training sets in the target language is available. Furthermore, this is evidence that multilingual datasets are crucial not only for the evaluation of multilingual dialogue systems, but also for better training them.  

Comparing different data scarcity regimes, the results in Tables~\ref{tab: multi source intent classification}--\ref{tab: multi source slot labelling} show that similar patterns emerge in few-shot performance on both tasks. However, in this case \textit{typ-closest} almost completely bridges the gap with \textit{multi-source}. This hints at the fact that even (larger) data from typologically diverse sources is no match for (smaller) in-domain supervision. Even more importantly, however, the performance of both transfer methods dramatically increases over their zero-shot counterparts.


The results of the study provide an interesting insight on cross-lingual NLU with single-source and multi-source training. In line with previous work \cite{ansell-etal-2021-mad-g}, they show that multi-source finetuning yields better performance than single-source, even when the single source is a language linguistically close to the target language. We assume that the reason is that larger linguistic diversity of training data leads to more generalised representations enabling the model to perform better on an unseen language. From a more practical perspective, with the limited annotation resources one should consider splitting it between more languages rather than annotating a large amount of data in one language only.

As a general finding, this small study suggests that: \textbf{1)} carefully picking source languages; \textbf{2)} transferring knowledge from multiple sources; and \textbf{3}) adapting models to few target examples may steer and substantially improve dialogue NLU performance in the future. All these aspects, in turn, ultimately depend on the ability to wisely balance the annotation budget across typologically diverse languages when creating multilingual datasets.

\subsection{Fluency of Generated Language: Rich Morphology and Code Switching}
\label{ssec:fluencygen}

Besides coping with a wide cross-lingual variability of user utterances in the NLU components, 
multilingual \tod systems also need to produce accurate \textit{and} fluent responses suited for the target language.
Given the dialogue history and constrained by the domain, the NLG module should output a response which is articulate, sounds native to the user, and fits in the given context, without breaking the flow of the multi-turn conversation \shortcite{garbacea2020neural}.

There are challenges which are common to both \tod and Natural Language Generation tasks in general, such as complex morphology (e.g., fusional languages like the Slavic genus, agglutinative languages such Finnish or Turkish, polysynthetic languages like Inuktitut). Generating fluent and grammatical text in such morphologically rich languages is naturally much more arduous than in isolating languages without inflectional and derivative morphology \shortcite<e.g., Vietnamese;>{kunchukuttan-etal-2014-shata,Gerz:2018emnlp}. This stems from the inability to hold all the possible word forms in the vocabulary with traditional word embeddings or from word ``over-segmentation'' with recent subword-based pretrained Transformers \shortcite{ronnqvist2019multilingual,Rust:2020arxiv}. NLG in morphologically rich languages can benefit from dedicated language-specific tokenizers \shortcite{Rust:2020arxiv}, the incorporation of linguistic features \shortcite{klemen2020enhancing}, or through multi-tasking, predicting word and morphological information simultaneously \shortcite{passban2018improving}.\footnote{Another problem in cross-lingual setups concerns the adaptation to different word orders, where NLG might be directly informed by typological information through structural interventions \shortcite{Ponti:2018acl} or re-ordering \shortcite{daiber2016universal}.}

Furthermore, there are linguistic phenomena specific to informal, colloquial language. For instance, code-switching is a phenomenon where interlocutors shift from one language to another during the conversation \shortcite{sankoff1981formal}. It has been shown that code-switching might even improve the chances of successful task completion and the system's perceived friendliness \shortcite{ahn-etal-2020-code}.
\shortciteA{banerjee2019graph} show that structure-aware generation is effective for code-switched texts, even when dependency parsers are not available. Furthermore, \shortciteA{khanuja2020gluecos} maintain that, to tailor cross-lingual models such as mBERT for code-switching settings, they should be \textit{fine-tuned} on code-switched data, since the lexical distribution in code-switched language is different from the union of two languages. Additionally, prior work demonstrate that \textit{pretraining} Transformer-based encoders on conversational data leads to significant improvements on all dialogue tasks \shortcite{henderson2019convert,mehri2020dialoglue}. Unfortunately, however, large code-switching dialogue corpora are not available yet. 

Language fluency and the more general user satisfaction, which concerns not only \textit{what} the system responds, but also \textit{how} it conveys information, cannot be entirely captured with fully automatic evaluation measures. This justifies the need to conduct human-centred evaluations in order to reliably trace any real improvements in the perceived user-friendliness and the general eloquence of \tod systems in different languages. This leads us to the next challenge, discussed in \S\ref{ss:eval}.

\subsection{Human-centred Evaluation of Multilingual \tod Systems}
\label{ss:eval}


A crucial compass to chart the development of \tod systems is evaluation \shortcite{deriu2021survey}. For the modular \tod pipeline, there are standard automated metrics to evaluate each component: accuracy and/or $F_1$ score for intent classification, $F_1$ score for slot labelling, or joint goal accuracy for DST. Recently, DialoGLUE \shortcite{mehri2020dialoglue}, a benchmark for \tod systems based on automated metrics, has been proposed. Unfortunately, DialoGLUE is available only in English. 
We thus hope that, similar to recent benchmarks for general-purpose (i.e., non-dialogue) cross-lingual NLU such as XTREME(-R) \shortcite{hu2020xtreme,Ruder:2021xtremer} and XGLUE \shortcite{Liang:2020xglue}, future work will strive to building comprehensive and community-supported multilingual \tod benchmarks. An additional aspect to consider when automatically evaluating multilingual ToD systems is the confusion between low cross- and multi-lingual performance and cultural variance. In other words, one needs to differentiate between the models performing badly due to their inability to support cross-lingual transfer linguistically versus their inability to transfer between different cultural contexts and realities \cite{hershcovich2022challenges}.


Evaluation of \tod systems still adds another layer of difficulty, typically not encountered in the above-mentioned benchmarks for general-purpose NLP tasks. In short, received wisdom cautions that strong performance on automated metrics does not always positively correlate with the overall user satisfaction with the system \shortcite{liu2016not,Smith:2022arxiv}. This means that human evaluation remains the most faithful way to evaluate the ultimate dialogue system usability \textit{and} usefulness.

Human evaluation for \tod systems aims to figure out whether the user was satisfied with the interaction (\textit{user satisfaction}) or whether the system has completed the task \shortcite<{\em task success}; >{deriu2021survey}. Generally, collecting human feedback is money- and time-consuming. Multilingual \tod systems inherit the same costs but also pose new questions for widespread human evaluation protocols. Firstly, the expenses for hiring qualified users scales linearly with the number of languages. Secondly, especially for lower-resource languages, sometimes it is hard (or impossible) to find fluent speakers on commonly used platforms such as Amazon Mechanical Turk. Thirdly, when hiring evaluators from different countries, one needs to consider whether cultural differences subsist which may alter the way user satisfaction is perceived or reported.

Another issue is the current lack of standardisation of human-based evaluation in multilingual ToD. \shortciteA{howcroft2020twenty} show that even the definitions of the terms `accuracy' and `fluency' for NLG differ between researchers, which makes the comparison between two evaluation runs virtually impossible. To this end, GENIE \cite{khashabi2021genie} and GEM \cite{gehrmann2021gem} were proposed, with the central aim of standardising protocols for more reproducible human evaluation. While creating such standardised protocols is feasible and more straightforward in a monolingual context, defining a unified protocol is still an open question in multilingual contexts. User satisfaction with a dialogue system is highly dependent on what the user sees as good, useful, appropriate versus bad, incorrect, pointless. These notions are highly culture-dependent \cite{kendall2005sociology}. It means that an appropriate response of a dialogue system in the Anglosphere can be wrong, impolite and even aggressive when ported without any adaptation to a different culture \cite{hershcovich2022challenges}. Thus, accounting for ethics, cross-cultural differences and social norms is crucial when evaluating multilingual NLP in general \cite{solaiman2021process} and dialogue systems in particular. A potential approach to make the evaluation protocols more informed of the culture is to involve social scientists into multilingual \tod research. They might provide guidance on how to ground the protocol in the target language culture, including taboo topics and sensitive discussion points that should be avoided by the systems.

\subsection{Voice-Based Multilingual Dialogue?}
\label{ss:speech}
This survey, following the current mainstream in monolingual and multilingual \tod modelling, has hitherto focused on text-based input. However, the assumption of working with clean, complete, and fluent input text may be na\"ive: in fact, it underestimates the errors cascading from imperfect automatic speech recognition (ASR) onto the subsequent text-based modules. While previous work on English \tod paid attention to recovering from ASR errors and incorporating imperfect ASR output into the text-based pipeline \shortcite[\textit{inter alia}]{henderson2014dstc2,mrkvsic2017neural,Ohmura:2018slt}, the crucial speech-to-text and text-to-speech bridges are typically overlooked in multilingual \tod research: this also means that the true abilities of voice-based \tod systems are likely misconstrued.

ASR and speech-to-text synthesis are wide research fields in their own right, also seeking an expansion towards multilinguality as a long-standing and crucial research goal \shortcite[\textit{inter alia}]{Le:2009speech,Ghoshal:2013icassp,Conneau:2020asr}. The current mainstream ASR paradigm for cross-lingual transfer also relies on large pretrained Transformer-based multilingual models \shortcite{Conneau:2020asr,Pratap:2020:50}. Similar to multilingual BERT or XLM-R in the text domain, a heavily parameterised ASR model is trained on a large multilingual corpus of pure speech, and then fine-tuned with smaller amounts of speech-to-text transcriptions in particular target languages \shortcite{Pratap:2020inter}. Nonetheless, even this approach, termed `massively multilingual' by \shortciteA{Pratap:2020:50}, spans only around 50 languages.\footnote{On top of this, recent research has shown that ASR does not provide equitable service to native speakers of the same language from different backgrounds \shortcite{Koenecke:2020pnas} or under different recording circumstances \shortcite{salesky2021sigtyp}.} Yet, cutting-edge fully unsupervised models, such as wav2vec-U \shortcite{baevski2021unsupervised}, hold promise to extend ASR even to languages without transcribed speech available. In particular, after the standard pre-training on unlabelled speech, wav2vec-U representations are fed to a Generative Adversarial Network, which tries to discriminate between real and generated phoneme sequences. The encouraging early results of this method on low-resource languages will need to be validated in future work, and its coupling with multilingual \tod is yet to come.

A similar situation is observed with multilingual text-to-speech (TTS) research: despite recent efforts, multilingual TTS modules are available for a tiny fraction of languages \shortcite{Zhang:2019inter,Nekvinda:2020inter}, even smaller than what multilingual ASR currently supports. However, breakthroughs in unsupervised TTS may be decisive to widen language coverage \shortcite{zhang2020unsupervised}.

This effectively means that voice-based \tod is currently out of reach for the overwhelming majority of the world's languages \shortcite{Joshi:2020acl}. More generally, in the pursuit of wider-scale and democratised \tod technology, we advocate an even tighter integration of speech-based and text-based modules in future work, as well as more realistic evaluation protocols which also include ASR and TTS error analyses. In fact, any future developments of multilingual \tod are also tightly coupled with parallel developments in multilingual ASR and TTS as standalone research areas.

\subsection{Other Related Areas} 
\label{ss:other}
We have attempted to cover multiple threads of research connected to multilingual conversational AI, as an extremely wide multi-disciplinary and multi-layered field. However, we also acknowledge that there are other areas related to the development and deployment of full-fledged and engaging multilingual \tod systems which remained out of our main focus. These other directions include (but are not limited to): \textbf{1)} making \tod systems more flexible and empathetic by relying on implicit conversational cues and (multilingual) emotion recognition \shortcite{pittermann2010emotion,heracleous2020integrating,meng2020dukenet}; \textbf{2)}  incorporating the information from miscellaneous knowledge bases to improve the system’s commonsense reasoning and world knowledge capabilities \shortcite{eric2017key,madotto2018end2endknowledge,haihong2019kb}; \textbf{3)} grounding dialogue in perceptual (typically visual) contexts \shortcite{DeVries:2017cvpr,AlAmri:2019cvpr,Shekhar:2019naacl,Agarwal:2020acl}.\footnote{Besides providing additional (situational) context to dialogues in general, multi-modal modelling might also be advantageous to multilingual settings. Indeed, visual input (e.g., images, videos) can also serve as a naturally occurring interlingua for cross-lingual alignment \shortcite{Kiela:2015emnlp,Gella:2017emnlp,Rotman:2018acl,Sigurdsson:2020cvpr}.} Stepping a bit further away, it is also quite reasonable to assume that further advances in machine translation for massively multilingual and low-resource settings \shortcite{Siddhant:2020acl,siddhant2020evaluating,Garcia:2020arxiv,Fan:2020arxiv} will also (continue to) have substantial impact on multilingual \tod by enhancing cross-lingual transfer capability.

%% file: 6-conclusion.tex
\section{Conclusion}
Enabling machines to converse similarly to humans is one of the central goals of AI. Achieving this in a multitude of the world's languages is an even more ambitious challenge. In this work, we have presented an overview of the current challenges and efforts, including the state-of-the-art methods and available datasets, and future directions concerning multilingual task-oriented dialogue (\tod) systems. 
In light of this survey, we can now attempt to answer the questions first posed in \S\ref{sec: introduciton} as follows:

\vspace{1.2mm}

\noindent (\textbf{A1}) Despite gaining much more traction recently, the availability of multilingual \tod \textbf{datasets} is still scarce, especially for NLG and end-to-end dialogue. While NLU is better documented, the datasets dedicated to this task often lack typological diversity in their language sample and idiomacity in their conversations (i.e., by avoiding `translationese' and cultural biases).
\vspace{1.1mm}

\noindent (\textbf{A2}) Given the paucity of multilingual \tod data, the most established \textbf{methods} include model transfer based on massively multilingual encoders and translation-based transfer (especially for NLG). Moreover, based on our experimental findings in \tod NLU, we identify multi-source transfer and few-shot fine-tuning as two promising solutions to substantially boost performance.
\vspace{1.1mm}

\noindent (\textbf{A3}) While it should remain aware of major (dis)similarities, multilingual \tod can and should borrow from other \textbf{related fields} of NLP the well-oiled `machinery' tried and tested to alleviate resource-poor settings. In particular, strategies to extend the language coverage of massively multilingual encoders, unsupervised domain adaptation, data augmentation, meta learning, and conditional generation may prove essential to extend the reach of multilingual \tod systems.
\vspace{1.1mm}

\noindent (\textbf{A4}) The \textbf{future challenges} of multilingual \tod include (but are not limited to): i) the generation of fluent text in morphologically complex languages and code-switching scenarios; ii) the refinement of strategies to collect expensive annotated data and enable human-centred evaluation protocols; iii) a tighter integration of speech-based and text-based modules, as recent unsupervised learning techniques hold promise to enable ASR and TTS for a plethora of (resource-poor) languages.

\vspace{1.2mm}
\noindent
With these findings, 
we aim to inspire more work in these important areas. In the long run, we hope that our overview will 
contribute to foster and guide future developments in \tod towards truly multilingual and inclusive conversational AI.

Finally, an additional contribution of this work, potentially useful to other researchers and practitioners interested in this emerging field, is an up-to-date online repository of all the datasets falling within the scope of multilingual \tod, which is available at \url{https://github.com/evgeniiaraz/datasets_multiling_dialogue}.

%% file: 7-appendix.tex
\appendix

\section{English NLU datasets}
\label{app: english nlu datasets}
\begin{table}[h!]
\centering
\resizebox{\textwidth}{!}{
\begin{tabular}{llccccc} 
\toprule
\textbf{Dataset}                                                                & \textbf{Task}                                                                       & \textbf{Language} & \textbf{Domains}                                                                                           & \textbf{Size} & \textbf{\# intents} & \textbf{\# slots}  \\ 
\midrule
\begin{tabular}[c]{@{}l@{}}Banking-77 \\ \shortcite{casanueva2020banking77}\end{tabular}    & intent classification                                                               & en                & banking                                                                                                    & 13083         & 77                  & N/A                \\ 
\hline
\begin{tabular}[c]{@{}l@{}}CLINC-150 \\ \shortcite{larson2019clinc150}\end{tabular}          & intent classification                                                               & en                & \begin{tabular}[c]{@{}c@{}}10 domains, \\ inter alia, \\ banking, work, \\ travel, small talk\end{tabular} & 23700         & 150                 & N/A                \\ 
\hline
\begin{tabular}[c]{@{}l@{}}HWU64 \\ \shortcite{liu2019hwu64}\end{tabular}                   & \begin{tabular}[c]{@{}l@{}}intent classification; \\ entity extraction\end{tabular} & en                & \begin{tabular}[c]{@{}c@{}}21 domains, \\ inter alia, \\ music, news, \\ calendar\end{tabular}             & 25716         & 64                  & 54                 \\ 
\hline
\begin{tabular}[c]{@{}l@{}}Restaurants-8K \\ \shortcite{coope2020restaurants8k}\end{tabular} & slot extraction                                                                     & en                & restaurant booking                                                                                         & 11929         & N/A                 & 5                  \\ 
\hline
\begin{tabular}[c]{@{}l@{}}Snips \\ \shortcite{coucke2018snips}\end{tabular}                & \begin{tabular}[c]{@{}l@{}}intent classification;\\ slot extraction\end{tabular}    & en                & \begin{tabular}[c]{@{}c@{}}7 domains, \\ inter alia, \\ music, weather, \\ restaurant\end{tabular}         & 14484         & 7                   & 39                 \\ 
\hline
\begin{tabular}[c]{@{}l@{}}ATIS\\ \shortcite{price1990atis}\end{tabular}                    & \begin{tabular}[c]{@{}l@{}}intent classification;\\ slot extraction\end{tabular}    & en                & airline travels                                                                                            & 5871          & 21                  & 120                \\
\bottomrule
\end{tabular}
}
\caption{English NLU datasets. This list is non exhaustive.}
\label{table:nlu english}
\end{table}

\section{English DST datasets}
\label{app: english dst datasets}
\begin{table}[h!]
\centering
\resizebox{\textwidth}{!}{
\begin{tabular}{llcccc} 
\toprule
\textbf{Dataset}                                                                       & \textbf{Task}                                                      & \textbf{Language} & \textbf{Domain}                                               & \begin{tabular}[c]{@{}c@{}}\textbf{Size}\\\textbf{ (dialogues)}\end{tabular} & \textbf{H2H / H2M}  \\ 
\midrule
\begin{tabular}[c]{@{}l@{}}DSTC1\\ \shortcite{raux2005letsgodataset,william2013dstc1}\end{tabular} & \begin{tabular}[c]{@{}l@{}}dialogue state\\ tracking\end{tabular}  & en                   & bus information                                               & 15886                                                                        & H2M                 \\ 
\hline
\begin{tabular}[c]{@{}l@{}}DSTC2\\ \shortcite{henderson2014dstc2}\end{tabular}                     & \begin{tabular}[c]{@{}l@{}}dialogue state \\ tracking\end{tabular} & en                   & \begin{tabular}[c]{@{}c@{}}restaurant \\ booking\end{tabular} & 3000                                                                         & H2M                 \\ 
\hline
\begin{tabular}[c]{@{}l@{}}WOZ2.0\\ \shortcite{wen2017network,mrkvsic2017neural}\end{tabular}      & \begin{tabular}[c]{@{}l@{}}dialogue state \\ tracking\end{tabular} & en                   & \begin{tabular}[c]{@{}c@{}}restaurant \\ booking\end{tabular} & 1200                                                                         & H2H                 \\
\bottomrule
\end{tabular}
}
\caption{Englsh DST datasets. This list is non exhaustive. Abbreviations: H2M -- human-to-machine; H2H -- human-to-human. }
\label{table: english  dst}
\end{table}

\newpage
\section{English end-to-end datasets}
\begin{table}[h!]
\centering
\resizebox{\textwidth}{!}{
\begin{tabular}{llcccc} 
\toprule
\textbf{Dataset}                                                            & \textbf{Task}                                                                                                                        & \textbf{Language} & \textbf{Domain}                                                                                   & \begin{tabular}[c]{@{}c@{}}\textbf{Size}\\\textbf{ (dialogues)}\end{tabular} & \textbf{Comments}                                                                                  \\ 
\midrule
\begin{tabular}[c]{@{}l@{}}MultiWOZ\\ \shortcite{budzianowski2018multiwoz}\end{tabular} & \begin{tabular}[c]{@{}l@{}}end-to-end;\\ dialogue state\\ tracking;\\ slot extraction;\end{tabular}                                  & en                   & \begin{tabular}[c]{@{}c@{}}7 domains, \\ including\\ restaurant, \\ taxi\end{tabular}             & 10438                                                                        & H2H;                                                                                               \\ 
\hline
\begin{tabular}[c]{@{}l@{}}Taskmaster-1\\ \shortcite{byrne2019taskmaster1}\end{tabular} & end-to-end                                                                                                                           & en                   & \begin{tabular}[c]{@{}c@{}}6 domains, \\ including\\ ordering pizza,\\ movie tickets\end{tabular} & 7708                                                                         & Self-dialogues;                                                                                    \\ 
\hline
\begin{tabular}[c]{@{}l@{}}MultiDoGo\\ \shortcite{peskov2019multidogo}\end{tabular}     & \begin{tabular}[c]{@{}l@{}}end-to-end;\\ intent \\ classification;\\ slot extraction;\\ dialogue acts\\ classification;\end{tabular} & en                   & \begin{tabular}[c]{@{}c@{}}6 domains,\\ including\\ airline, \\ software\end{tabular}             & 40576                                                                        & H2H                                                                                                \\ 
\hline
\begin{tabular}[c]{@{}l@{}}ConvAI2\\ \shortcite{dinan2019convai2}\end{tabular}          & end-to-end                                                                                                                           & en                   & \begin{tabular}[c]{@{}c@{}}chit chat, \\ not goal oriented\end{tabular}                           & 19893                                                                        & \begin{tabular}[c]{@{}c@{}}H2H;\\ derived from \\Persona-Chat\\ \shortcite{zhang2018personachat}\end{tabular}  \\
\bottomrule
\end{tabular}
}
\caption{English datasets for end-to-end training. H2H means human-to-human dialogues.}
\label{table: english e2e}
\end{table}

\section{Multilingual chit-chat datasets}
\begin{table}[!h]
\centering
\resizebox{\textwidth}{!}{
{\footnotesize
\begin{tabular}{llcccc} 
\toprule
\textbf{Dataset} & \textbf{Task} & \textbf{Language(s)} & \textbf{Domain}                                                                                     & \begin{tabular}[c]{@{}c@{}}\textbf{Size}\\\textbf{ (dialogues)}\end{tabular}                                                     & \textbf{Comments}                                                                                                                                    \\ 
\midrule
\rowcolor{Gray}
\multicolumn{6}{l}{\sc Monolingual datasets}                                                                                                                                                                                                                                                                                                                                                                                                                                                                                                                                                                                              \\ 
\midrule
\begin{tabular}[c]{@{}l@{}}DuConv\\ \shortcite{wu-etal-2017-sequential}\end{tabular} & end-to-end                                                                       & zh                                                                       & chit-chat                                                                                           & \begin{tabular}[c]{@{}c@{}}1060000\\ (context-response\\ pairs)\end{tabular}                                                     & \begin{tabular}[c]{@{}c@{}}H2H; web \\ scraped from \\ social network;\end{tabular}                                                                  \\ 
\hline
\begin{tabular}[c]{@{}l@{}}KdConv\\ \shortcite{zhou2020kdconv}\end{tabular}           & end-to-end                                                                       & zh                                                                       & \begin{tabular}[c]{@{}c@{}}chit-chat about\\ films. music,\\ travel\end{tabular}                    & 4500                                                                                                                             & \begin{tabular}[c]{@{}c@{}}H2H; using\\ an external \\ knowledge base;\end{tabular}                                                                  \\ 
\midrule
\rowcolor{Gray}
\multicolumn{6}{l}{\sc Multilingual datasets}                                                                                                                                                                                                                                                                                                                                                                                                                                                                                                                                                                                              \\ 
\midrule
\begin{tabular}[c]{@{}l@{}}XPersona\\ \shortcite{lin2020xpersona}\end{tabular}       & end-to-end                                                                       & \begin{tabular}[c]{@{}c@{}}en, it, fr, \\ id, zh, ko, \\ ja\end{tabular} & \begin{tabular}[c]{@{}c@{}}chit-chat \\ (persona chats)\end{tabular}                                & \begin{tabular}[c]{@{}c@{}}19893 [en]\\ 17158 [it]\\ 17375 [fr]\\ 17846 [id]\\ 17322 [zh]\\ 17477 [ko]\\ 17428 [ja]\end{tabular} & \begin{tabular}[c]{@{}c@{}}H2H;\\ automatically \\ translated from \\ \shortcite{dinan2019convai2};\end{tabular}                                                 \\
\bottomrule
\end{tabular}
}%
}
\caption{Multilingual chit-chat datasets for end-to-end training. H2H means human-to-human dialogues.}
\label{table: multi e2e chitchat}
\end{table} 

\section{Typologically closest source language for each target language in MultiATIS++}

\begin{table}[!h]
\centering
\begin{tabular}{lccccccccc} 
\toprule
target langugae                                                                 & de & en & es & fr & hi & ja & pt & tr & zh  \\ \cmidrule(lr){2-10}
\begin{tabular}[c]{@{}l@{}}linguistically closest\\source language\end{tabular} & en & de & pt & pt & zh & zh & es & de & ja  \\
\bottomrule
\end{tabular}
\end{table}